%% file: main.tex
\documentclass{article}

\PassOptionsToPackage{numbers, compress}{natbib}

\usepackage[main, final]{neurips_2026}

\usepackage[utf8]{inputenc} 
\usepackage[T1]{fontenc}    
\usepackage{hyperref}       
\usepackage{url}            
\usepackage{booktabs}       
\usepackage{amsfonts}       
\usepackage{amsmath}        
\usepackage{nicefrac}       
\usepackage{microtype}      
\usepackage{xcolor}         

\usepackage{multirow}
\usepackage{makecell}
\usepackage{graphicx}
\usepackage{caption}
\usepackage{subcaption}
\usepackage{svg}
\usepackage{booktabs}
\usepackage{tabularx}
\usepackage{makecell}
\usepackage{array}
\usepackage{enumitem}
\usepackage{wrapfig}
\usepackage{needspace}
\usepackage{float}

\title{Dominant-Layer ZO: A Single Layer Dominates Zeroth-Order Fine-Tuning of LLMs}

\author{%
\textbf{Wanhao Yu}$^{1}$ \quad
\textbf{Ziyan Wang}$^{1}$ \quad
\textbf{Zheng Wang}$^{2}$ \quad
\textbf{Abeer Matar Almalky}$^{3}$ \\
\textbf{Yihang Zuo}$^{4}$ \quad
\textbf{Shuteng Niu}$^{5}$ \quad
\textbf{Sen Lin}$^{2}$ \quad
\textbf{Adnan Siraj Rakin}$^{3}$ \\
\textbf{Deliang Fan}$^{4}$ \quad
\textbf{Li Yang}$^{1\dagger}$ \\
\\[-0.4em]
$^{1}$University of North Carolina at Charlotte \quad
$^{2}$University of Houston \\
$^{3}$State University of New York at Binghamton \quad
$^{4}$Arizona State University \\
$^{5}$Texas Christian University
}

\begin{document}

\maketitle
\begingroup
\renewcommand{\thefootnote}{\ensuremath{\dagger}}
\footnotetext[2]{Corresponding author.}
\endgroup

\begin{abstract}
Zeroth-order (ZO) optimization enables memory-efficient fine-tuning of large language models (LLMs) using only forward passes, but it remains unclear how useful adaptation is distributed across layers. In this work, we reveal a surprising phenomenon: ZO fine-tuning is sharply dominated by a single decoding layer. Across multiple LLM families and downstream tasks, fine-tuning this dominant layer alone consistently matches or even exceeds full-model ZO fine-tuning. We further show that the dominant layer is task-agnostic but model-specific, and can be identified before training through a simple inference-only analysis of activation outliers. Specifically, the dominant layer consistently aligns with the first activation-outlier layer in the pre-trained model. To explain this phenomenon, we analyze how perturbation effects propagate under ZO optimization. We find that the dominant layer combines two key properties: high perturbation sensitivity and early placement in the residual stream, allowing perturbation-induced effects to propagate and accumulate through remaining subsequent decoding layers. As a result, this layer produces stronger optimization signals under forward-only updates. Extensive experiments on LLaMA2-7B and Qwen3-8B across nine benchmarks show that dominant-layer ZO fine-tuning improves average performance over full-model MeZO and LoRA-based ZO fine-tuning while achieving up to 4.52$\times$ training speedup. Code is available at \url{https://github.com/uncc-efficient-ai/dominant-layer-zo}.

\end{abstract}

\input{sections/intro}
\input{sections/related_work}

\input{sections/main_results}

\input{sections/experiments}

\input{sections/conclusion}
\clearpage
\input{sections/acknowledgments}
\bibliographystyle{plainnat}
\bibliography{sections/references}

\newpage
\appendix
\input{sections/appendix}

\end{document}

%% file: sections/intro.tex
\section{Introduction}

Zeroth-order (ZO) optimization has recently emerged as a promising approach for memory-efficient fine-tuning of large language models (LLMs) \citep{zhang2024revisiting}. Instead of computing first-order (FO) gradients through backpropagation, ZO methods estimate update directions using only forward evaluations, typically by measuring loss differences under random parameter perturbations \citep{spall1992multivariate}. Building on this idea, MeZO shows that pre-trained LLMs can be fine-tuned with near inference-level memory \citep{malladi2023fine}. Subsequent studies further improve convergence and accuracy by reducing the variance of ZO gradient estimates through sparse parameter perturbation \citep{liu2024sparse, guo2025staticzo},  low-rank or structured perturbation spaces \citep{chen2025lozo, lin2026agzo}, and more stable or faster optimizer designs \citep{chen2019zoadamm,dang2025fzoofastzerothorderoptimizer}. Despite these advances, existing methods largely treat ZO fine-tuning as a full-model process, without explaining how useful adaptation differs across layers. This leaves a fundamental question unanswered: \textit{under forward-only updates, where does useful optimization actually occur inside LLM architectures?}

In this work, we reveal an unexpected phenomenon: \textbf{useful ZO adaptation is not spread broadly across layers, but is dominated by a single layer.} 

To study this phenomenon, we first conduct a systematic layer-wise analysis across multiple LLMs and downstream tasks, where we fine-tune one layer at a time under identical ZO updates while freezing all other layers. The results show a highly uneven layer-wise pattern: most layers provide little or no improvement over the no fine-tuning baseline, while a single layer consistently achieves performance comparable to, or even exceeding, full-model ZO fine-tuning. We refer to this layer as \textbf{\textit{dominant layer}}. Moreover, the dominant layer is \textit{task-agnostic but model-specific}: for a given LLM, the same layer consistently dominates across the evaluated tasks, while different model families may have different dominant-layer indices. In contrast, under matched first-order gradient fine-tuning, improvements are more evenly spread across layers, and no single layer consistently dominates. 
This contrast shows that the dominant-layer phenomenon is unique for ZO optimization, which follows a different layer-wise adaptation pattern from first-order (FO) fine-tuning.

We further study how to efficiently identify this dominant layer without expensive layer-wise ZO fine-tuning. Inspired by the known \textit{activation outlier} phenomenon in LLMs \citep{dettmers2022gpt3,xiao2023smoothquant}, where a small number of activations exhibit extremely large magnitudes at specific dimension indices across layers in an input-independent manner \citep{sun2024massive,an2025systematic}, we find that the dominant layer aligns with the first layer where activation outliers emerge. Based on this observation, we design a simple inference-only selection method: given a small calibration set, we run the pre-trained LLM forward, measure layer-wise activation statistics, and select the first layer that shows a clear outlier pattern. This method avoids exhaustive layer-wise ZO fine-tuning and identifies the dominant layer before training begins.

Finally, we explain why this dominant layer emerges under ZO fine-tuning. Unlike first-order optimization, ZO estimates updates only from final-loss differences caused by random perturbations. Therefore, a layer can contribute more to ZO fine-tuning when its perturbation has a stronger effect on the forward computation. We find that the dominant layer satisfies this condition because it appears early in the model and aligns with the first activation-outlier layer. Perturbations at this layer enter the residual stream and affect the activations of all remaining layers. This propagation allows the perturbation effect to be repeatedly transformed and accumulated before reaching the final loss, leading to larger final-loss changes and a stronger loss-difference signal for ZO updates.

This finding of a dominant layer in ZO fine-tuning has both practical and conceptual implications. Practically, this finding suggests that updating only a single layer can be sufficient to achieve performance comparable to full-model MeZO. More importantly, we hope this finding provides insight for future ZO method design, such as explicitly considering where useful updates arise across layers or making updates to non-dominant layers more effective.

Our contributions can be summarized as follows:
\begin{itemize}[leftmargin=*]
    \item We discover a dominant-layer phenomenon in ZO fine-tuning: tuning a single layer can recover, and sometimes exceed, the performance of full-model ZO fine-tuning.
    \item We show that the dominant layer is task-agnostic but model-specific, and can be efficiently identified before training using the first activation-outlier layer. 
    \item We explain why the dominant layer learns well under ZO fine-tuning:  residual connection propagation amplifies its perturbation effect, leading to larger final-loss changes and stronger ZO update signals.
    \item We validate the dominant-layer ZO fine-tuning through extensive experiments on two LLMs, LLaMA2-7B and Qwen3-8B, across nine downstream tasks. Compared to MeZO, dominant-layer ZO fine-tuning improves the average score by $0.86\%$ over full-model and $0.61\%$ over LoRA-based \citep{hu2022lora} ZO fine-tuning. In addition, it achieves a 1.12$\times$$-$4.52$\times$ speedup in ZO fine-tuning relative to full-model MeZO.

\end{itemize}

%% file: sections/related_work.tex
\section{Related Work}

\paragraph{Zeroth-order LLM Fine-tuning.}
Zeroth-order (ZO) optimization estimates update directions from function values rather than explicit backpropagated gradients, using methods such as SPSA, forming its classical foundation \citep{spall1992multivariate,nesterov2017random}. Recently, MeZO \citep{malladi2023fine} first shows that LLMs can be fine-tuned for downstream tasks with inference-level memory, making ZO a memory-efficient alternative to backpropagation for large models. In practice, MeZO estimates gradients by applying random perturbations to model parameters and measuring the loss difference between two forward passes, without storing intermediate activations for backward propagation. Beyond memory-efficient fine-tuning, forward-only and ZO optimization have also been explored for mitigating catastrophic forgetting in continual learning~\citep{feng2025zeroflow,yu2026memory}. 

To reduce gradient-estimation variance and accelerate convergence for more accurate and efficient fine-tuning, follow-up work mainly improves ZO fine-tuning along three directions. First, one line of work reduces the trainable or perturbed parameter scope through sparse parameter selection \citep{liu2024sparse}, transferable static sparsity \citep{guo2025staticzo}, or random layer-wise sparse updates \citep{wang2024lezo}. Second, another line of work reduces gradient-estimation variance by designing more informative perturbation directions, including low-rank directions in LOZO \citep{chen2025lozo}, random subspaces in SubZero \citep{yu2025subzero}, activation-derived directions in AGZO \citep{lin2026agzo}, and curvature-aware directions in HiZOO \citep{zhao2024hizoo}. Third, optimizer-level methods modify the update rule to improve optimization speed and stability, including layer-wise calibration in DiZO \citep{tan2025harmony}, clipping and annealing in HELENE \citep{zhao2024helene}, faster estimators in FZOO \citep{dang2025fzoofastzerothorderoptimizer}, and learned update rules in ZO Fine-tuner \citep{zhang2025learning}. AGZO~\citep{lin2026agzo} uses activations to guide perturbation directions, while \citet{guo2025staticzo} use first-order gradients and Fisher information to select a sparse parameter subset. Our work studies where effective ZO adaptation occurs across layers and identifies a dominant layer using only inference-time activation statistics.

\paragraph{Selective Layer-wise Fine-tuning.}
Layer-wise fine-tuning and layer-importance analysis have been widely studied in first-order LLM adaptation. Parameter-efficient tuning methods such as adapters and LoRA \citep{houlsby2019parameter,hu2022lora}, together with intrinsic-dimensionality analyses \citep{aghajanyan2021intrinsic}, suggest that effective adaptation often lies in a smaller update space than full-model fine-tuning implies. Recent methods further exploit layer-wise importance: LISA selectively freezes middle layers \citep{pan2024lisa}, ILA identifies layers critical for alignment \citep{shi2025understanding}, and IST/OwLore update selected layers based on layer importance or outlier-weighted sampling \citep{yao2024layer,li2025outlier}. In contrast, to the best of our knowledge, our work is the first to systematically analyze how layer-wise fine-tuning behaves under ZO optimization.

\paragraph{Outlier Activations in LLMs.}
Outlier activation is a common phenomenon in LLMs, first highlighted by LLM.int8()~\citep{dettmers2022gpt3} as a unique challenge for model compression: a small number of activation dimensions exhibit extremely large magnitudes compared to the average of the activation distribution. One important property is that these outliers consistently appear in the same activation dimensions across different inputs, suggesting that they come from the model structure rather than from any specific input sample~\citep{sun2024massive, an2025systematic}. Following this observation, a series of works study how to address activation outliers for efficient model compression, especially for quantization and pruning. For example, LLM.int8() isolates outlier features for mixed-precision inference \citep{dettmers2022gpt3}, while SmoothQuant migrates activation outlier difficulty into weights to enable accurate low-bit quantization \citep{xiao2023smoothquant}.


%% file: sections/main_results.tex
\section{Dominant Layer in ZO Fine-Tuning: Discovery and Identification}
\subsection{Preliminary: Zeroth-Order Optimization}
Following the classical two-point SPSA estimator \citep{spall1992multivariate} adopted by MeZO for ZO fine-tuning of LLMs \citep{malladi2023fine}, we estimate gradients from two perturbed loss evaluations. At iteration $t$ with  parameters $\theta_t$ and minibatch $\mathcal{B}_t$,  we sample random perturbation vector $z_t$ and compute 
the ZO gradient estimate as:
\begin{equation}
\widehat{g}_t =
\frac{
\mathcal{L}(\theta_t + \epsilon z_t;\mathcal{B}_t)
-
\mathcal{L}(\theta_t - \epsilon z_t;\mathcal{B}_t)
}{2\epsilon} z_t,
\label{eq:zo_grad}
\end{equation}
and the parameter update is
\begin{equation}
\theta_{t+1} = \theta_t - \eta_t \widehat{g}_t,
\label{eq:zo_update}
\end{equation}
where $\epsilon$ is the perturbation scale and $\eta_t$ is the learning rate. 

\subsection{Empirical Discovery: A \textit{Dominant Layer} Exists in ZO Fine-Tuning}
\label{sec:3.2}
We begin by analyzing how ZO fine-tuning behaves across layers by isolating each layer’s contribution. Specifically, based on MeZO, we fine-tune one layer at a time while freezing all other layers, using the same ZO update configuration for every layer. We conduct this study on LLaMA2-7B \citep{touvron2023llama} across multiple tasks, including WSC \citep{levesque2012winograd}, COPA \citep{roemmele2011copa}, and DROP \citep{dua2019drop}, which cover classification, multiple-choice, and generation settings.

\begin{figure}[t]
    \centering
    \begin{subfigure}[t]{0.485\columnwidth}
        \centering
        \includegraphics[width=\linewidth]{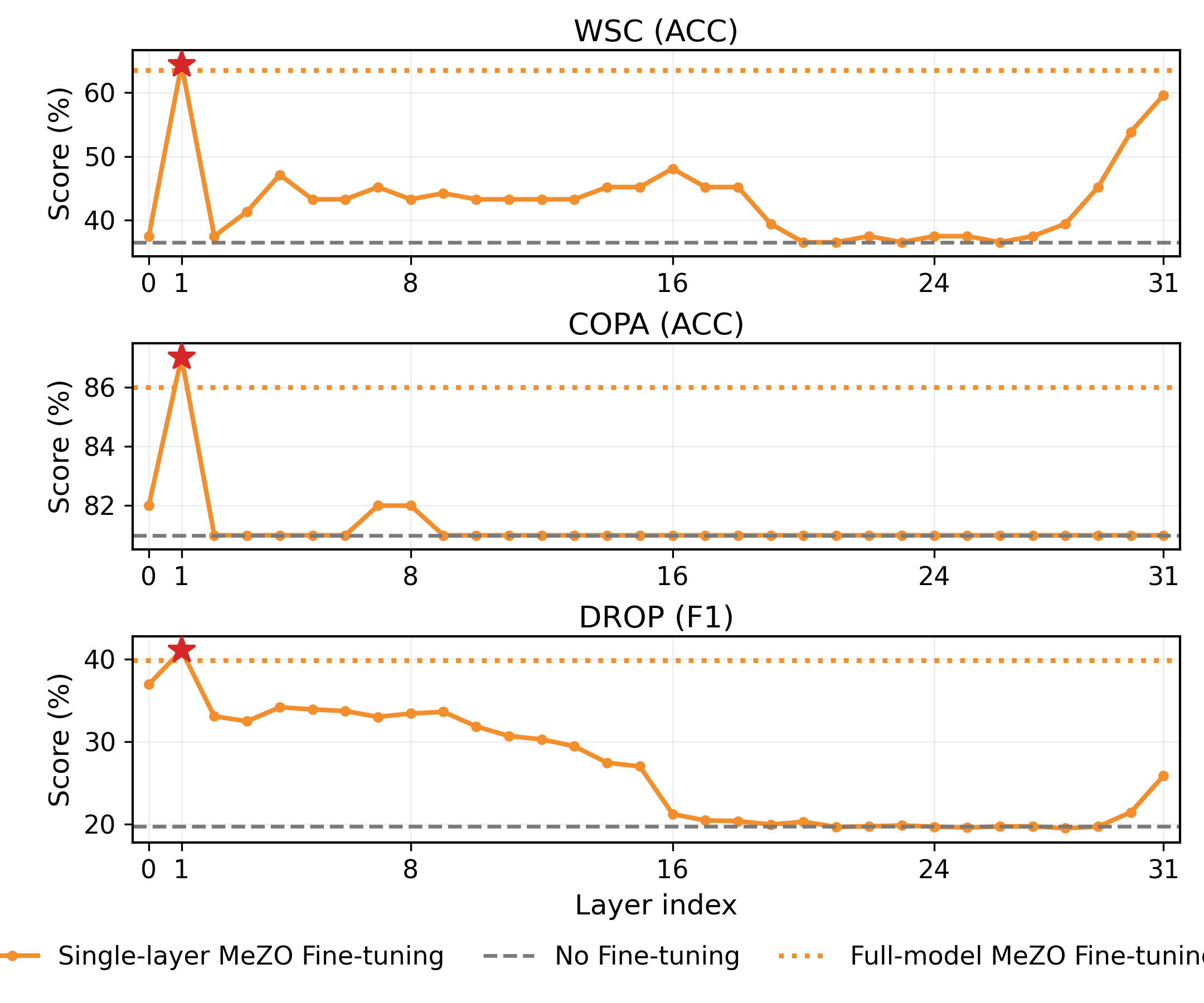}
        \caption{MeZO layerwise results.}
        \label{fig:mezo_layerwise_three_datasets}
    \end{subfigure}
    \hfill
    \begin{subfigure}[t]{0.485\columnwidth}
        \centering
        \includegraphics[width=\linewidth]{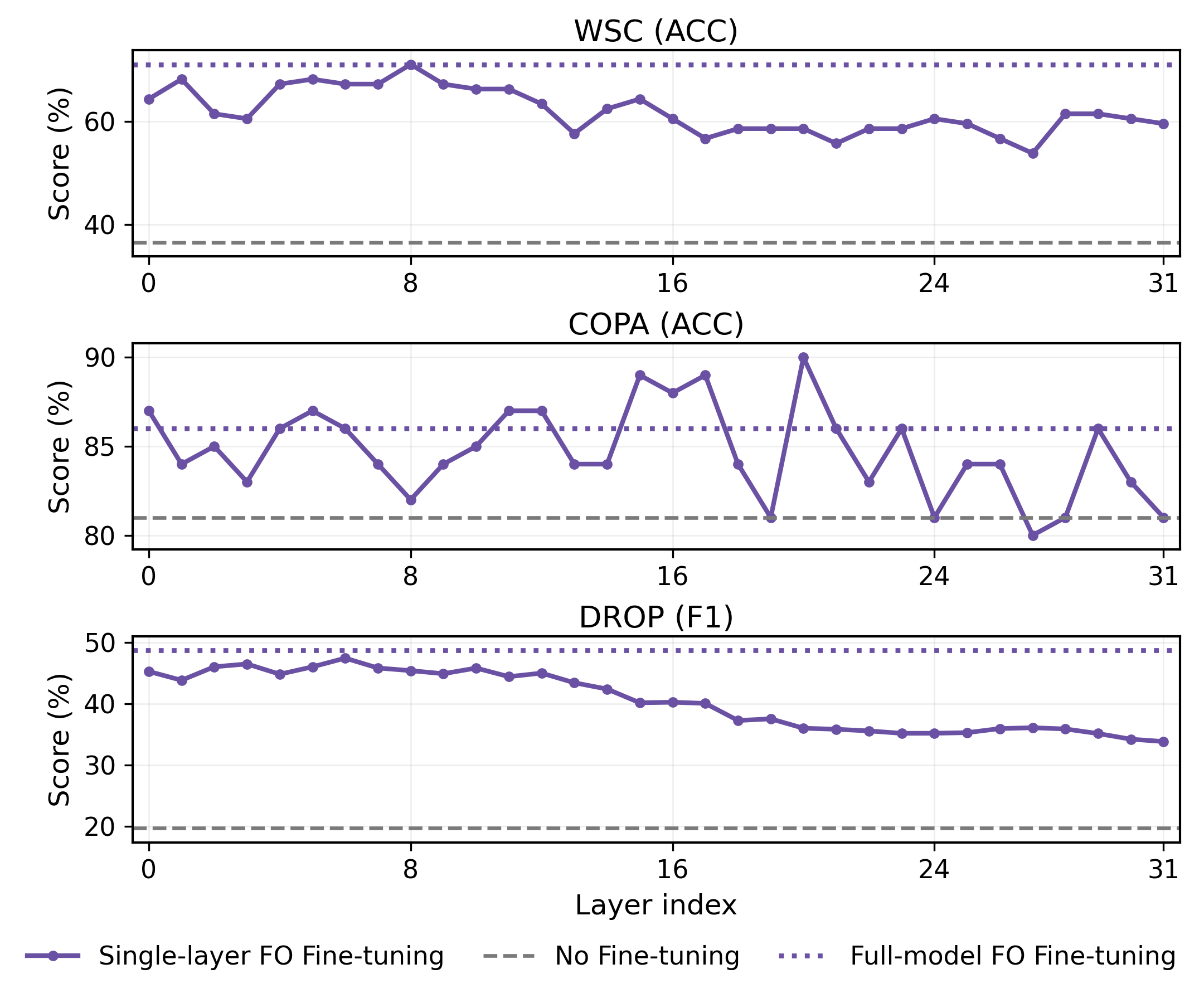}
        \caption{FO layerwise results.}
        \label{fig:fo_layerwise_three_datasets}
    \end{subfigure}
    \caption{Layer-wise fine-tuning on LLaMA2-7B results across three representative datasets.}
    \label{fig:fo_vs_mezo_layerwise_side_by_side}
\end{figure}

As shown in Figure~\ref{fig:fo_vs_mezo_layerwise_side_by_side}, we make two key observations:

\noindent\textbf{(1) ZO fine-tuning is highly uneven across layers.} Performance varies significantly across layers, and most layers provide little or no improvement over the no fine-tuning baseline. For example, on the COPA dataset, only a small subset of layers (4 out of 32 in LLaMA2-7B) improve accuracy after fine-tuning, while the majority remain close to the baseline.

\noindent\textbf{(2) A \textit{dominant layer} clearly emerges in ZO fine-tuning.} One specific layer achieves substantially higher performance than other layers and matches, or even exceeds, full-model MeZO fine-tuning. We refer to this layer as the \textbf{dominant layer}.
Moreover, we find that this layer has two important properties. First, it is \textbf{task-agnostic}: for a given LLM model, the same layer consistently dominates across different tasks. For example, in LLaMA2-7B, layer 1 achieves the best performance on all three tasks. Second, it appears to be \textbf{model-specific}: different model families may have different dominant-layer indices. For example, the dominant layer is layer 1 in LLaMA2-7B and layer 6 in Qwen3-8B, with the Qwen3-8B layer-wise analysis provided in Appendix~\ref{app:layer_sweeps}.





To further examine whether this behavior is specific to ZO, we repeat the same layer-wise analysis using first-order fine-tuning. As shown in Figure~\ref{fig:fo_layerwise_three_datasets}, first-order fine-tuning exhibits a different pattern: most layers achieve clear improvements over the baseline, and no single layer consistently dominates. This contrast indicates that fine-tuning effectiveness is more concentrated across layers under ZO than under FO in these experiments.

\begin{figure}[htbp]
    \centering
    \begin{subfigure}[t]{0.243\textwidth}
        \centering
        \includegraphics[width=\linewidth]{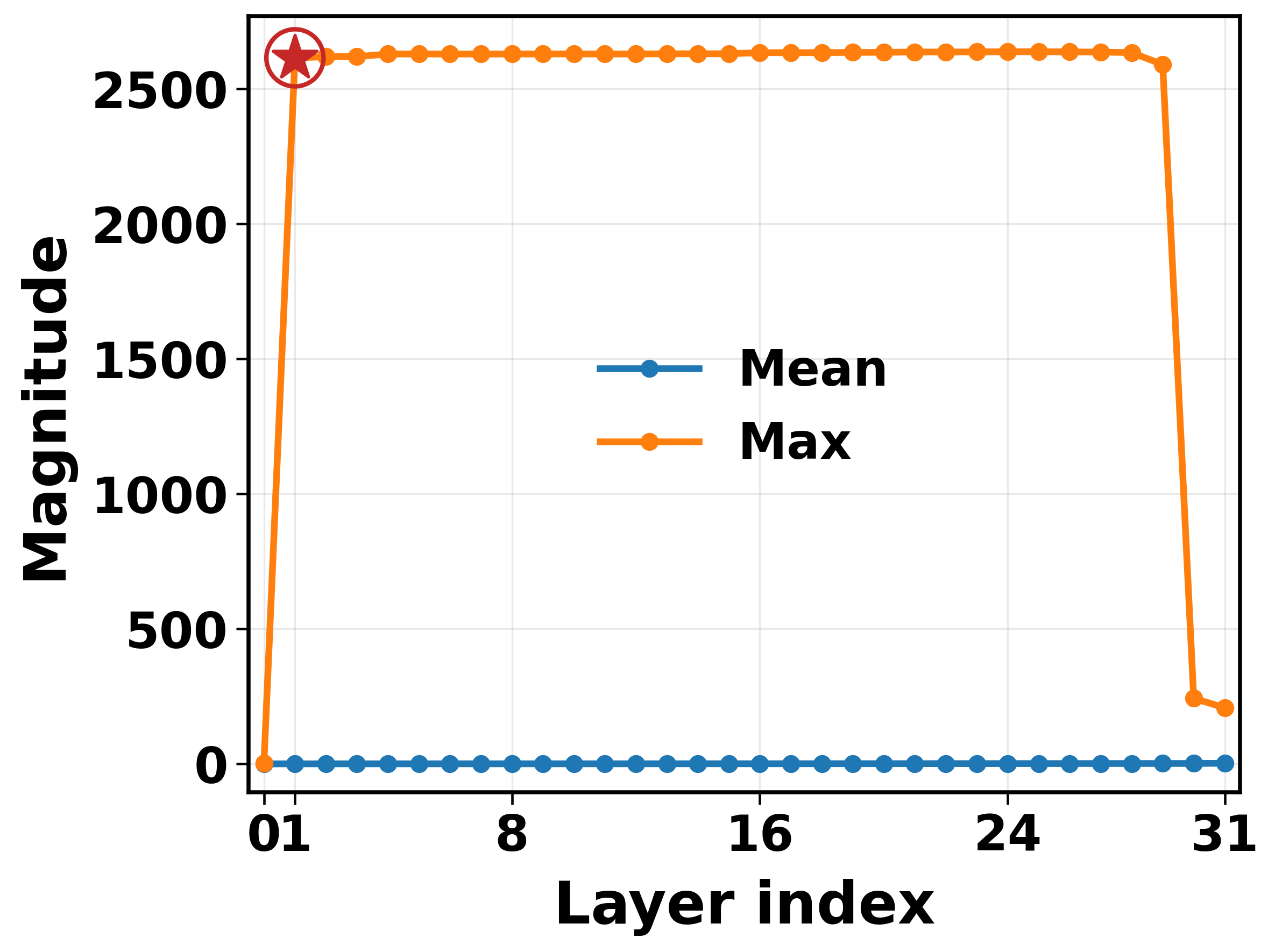}
        \caption{LLaMA2-7B on COPA.}
        \label{fig:llama2_copa_activation}
    \end{subfigure}\hfill
    \begin{subfigure}[t]{0.243\textwidth}
        \centering
        \includegraphics[width=\linewidth]{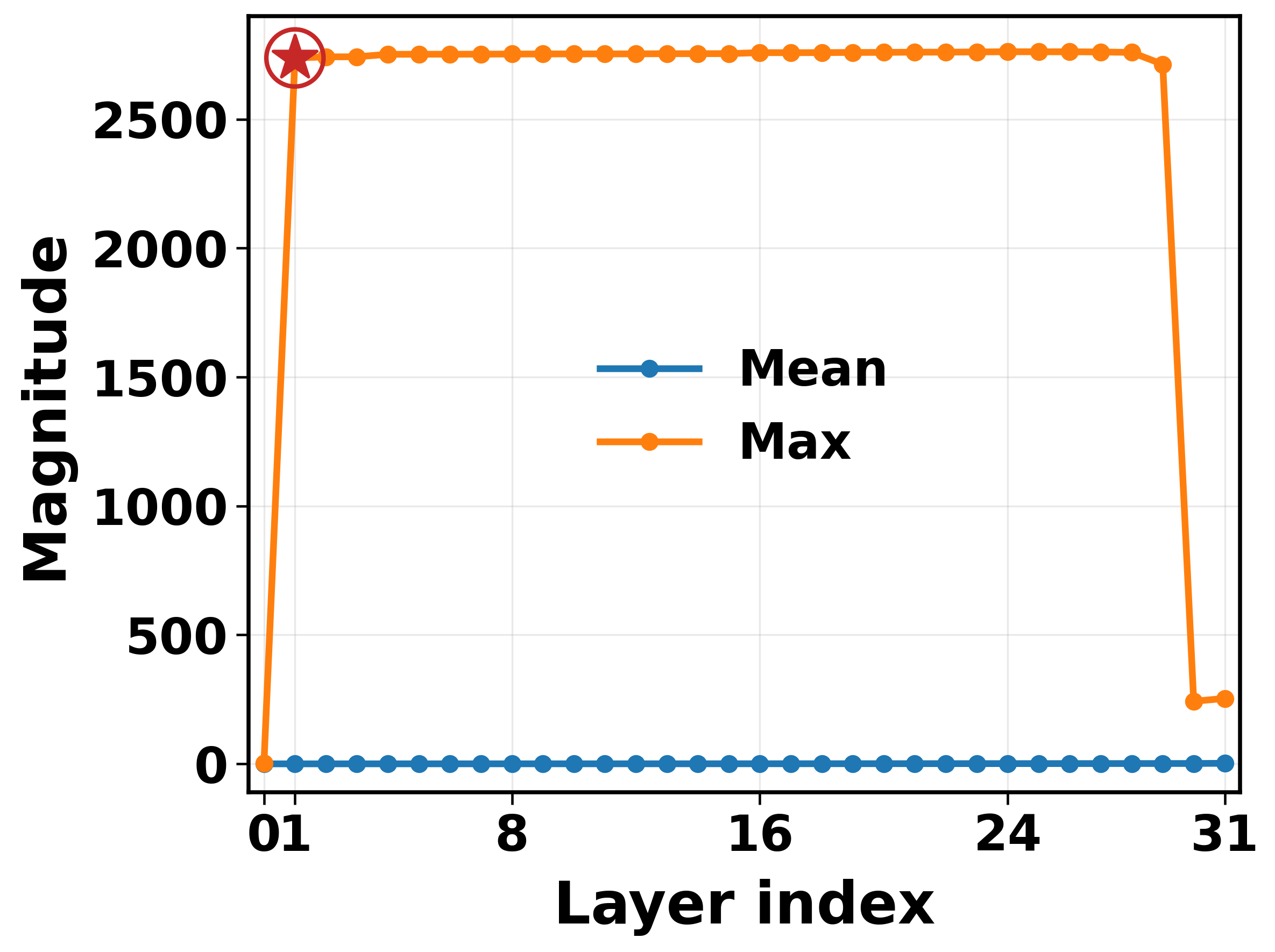}
        \caption{LLaMA2-7B on WSC.}
        \label{fig:llama2_wsc_activation}
    \end{subfigure}\hfill
    \begin{subfigure}[t]{0.243\textwidth}
        \centering
        \includegraphics[width=\linewidth]{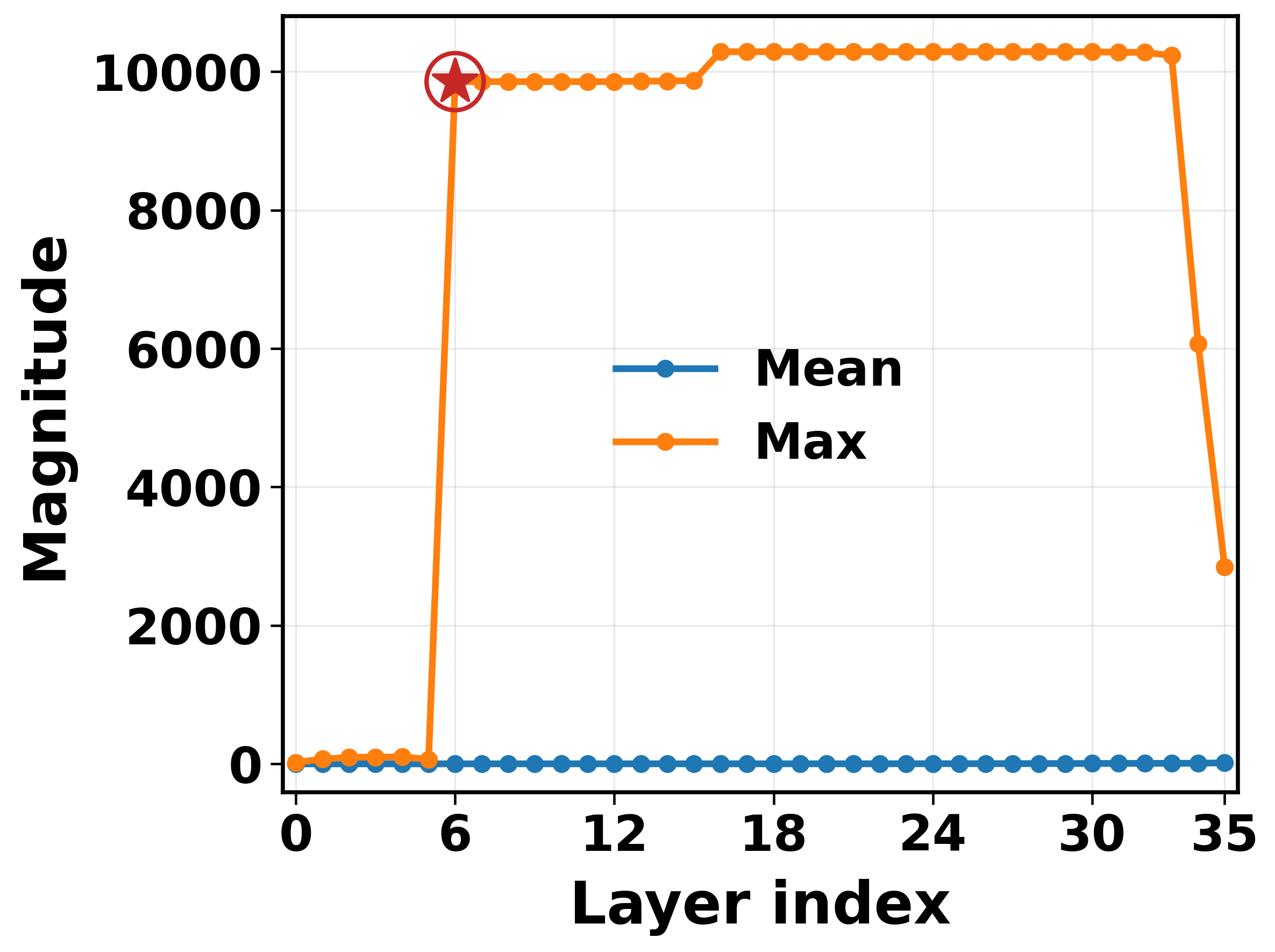}
        \caption{Qwen3-8B on COPA.}
        \label{fig:qwen3_copa_activation}
    \end{subfigure}\hfill
    \begin{subfigure}[t]{0.243\textwidth}
        \centering
        \includegraphics[width=\linewidth]{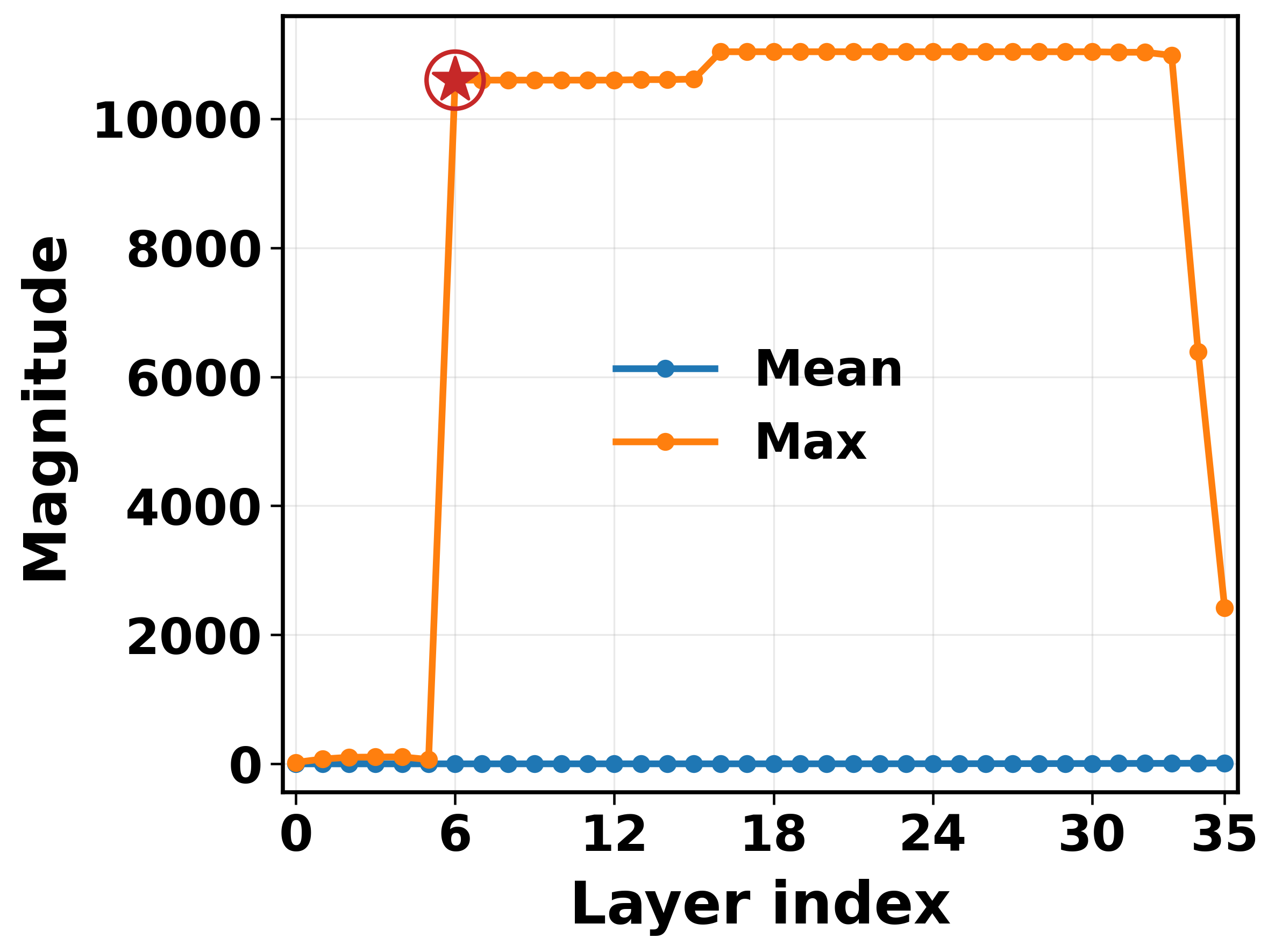}
        \caption{Qwen3-8B on WSC.}
        \label{fig:qwen3_wsc_activation}
    \end{subfigure}
    
    \caption{
    Mean and Maximum of Layer-wise output activation magnitudes for LLaMA2-7B and Qwen3-8B on COPA and WSC training samples. The highlighted point indicates the first activation-outlier layer, where the maximum output activation magnitude shows a clear jump.
    }
    \label{fig:activation_magnitude_four_plots}
\end{figure}




\subsection{Identifying the \textit{Dominant Layer} via Activation Outliers}
\label{sec:layer_identification}

The layer-wise analysis above identifies the dominant layer, but it requires fine-tuning each layer separately, which is computationally expensive. We therefore ask whether the dominant layer can be identified before fine-tuning, using only lightweight signals from the pre-trained model.

Our key observation is that the dominant layer aligns with the first layer where activation outliers emerge. This is motivated by the activation-outlier phenomenon: activation outliers appear consistently at specific activation dimensions and are agnostic to input data, which exhibits a similar property to the dominant layer observed in Section~\ref{sec:3.2}. As shown in Figure~\ref{fig:activation_magnitude_four_plots}, the first activation-outlier layer coincides with the dominant layer identified by layer-wise ZO fine-tuning. For example, the first outlier layer appears at layer 1 in LLaMA2-7B and layer 6 in Qwen3-8B, matching the dominant layers identified by layer-wise analysis.


Based on this observation, we propose a simple inference-only selection method. Given a small calibration set, we run the pre-trained model forward and compute layer-wise activation statistics. Following the activation-outlier criterion of \citet{an2025systematic}, we select the earliest layer whose maximum-to-mean activation magnitude ratio exceeds 1000. This method identifies the dominant layer before training and avoids expensive layer-wise ZO fine-tuning.






\section{Why Does the \textit{Dominant Layer} Emerge in ZO Fine-Tuning?}
\label{sec:mechanism}

\begin{figure*}[h]
    \centering
    \includegraphics[width=\linewidth]{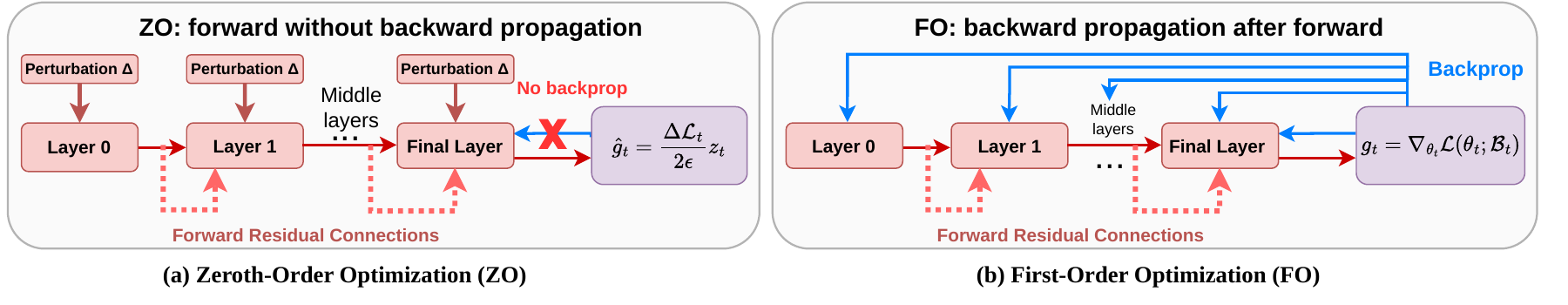}
    \caption{Comparison of optimization signal flow in ZO and FO. In ZO, gradients are estimated only from forward. In FO, backpropagation provides exact gradients throughout the network.}
    \label{fig:zo_fo_signal_flow}
\end{figure*}

\begin{wrapfigure}{r}{0.45\textwidth}
  \centering
  \vspace{-14pt}
  \includegraphics[width=\linewidth]{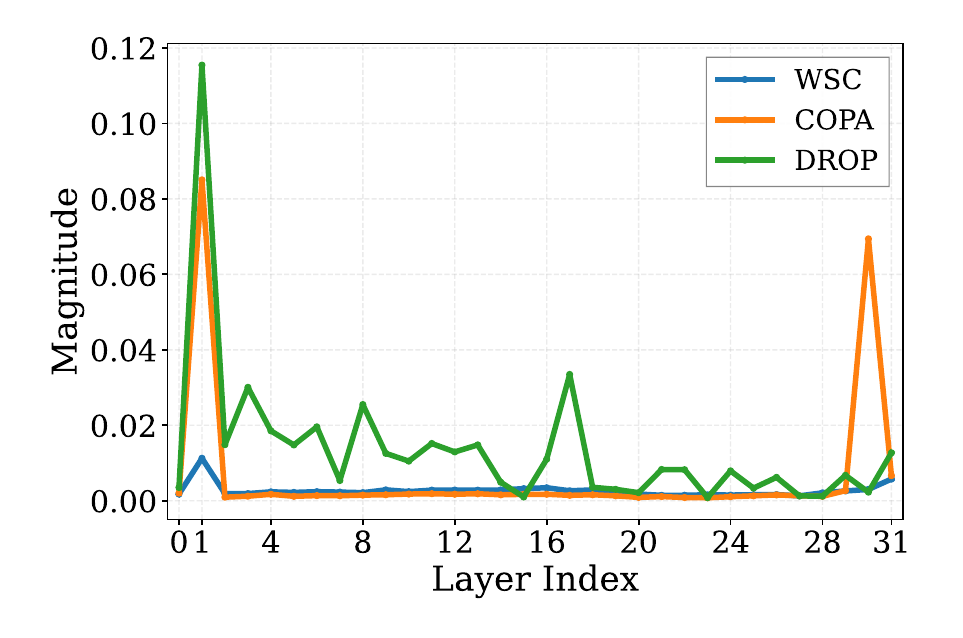}
\caption{Layerwise Loss Change under 1-step Perturbation on LLaMA2-7B across 3 tasks.}
\label{fig:perturbation_loss_change}
\end{wrapfigure}
ZO fine-tuning relies entirely on changes in the final loss under parameter perturbations. As a result, layers differ in effectiveness based on how strongly their perturbations influence the final loss, as illustrated in Figure~\ref{fig:zo_fo_signal_flow}(a). In contrast, FO fine-tuning distributes gradients across layers through backpropagation, preventing such concentration, as shown in Figure~\ref{fig:zo_fo_signal_flow}(b). To evaluate how each layer contributes to the final loss change, we perform a perturbation-only sensitivity analysis before ZO fine-tuning, where we apply random perturbations to one layer at a time and measure the resulting change in final loss. Here, perturbation sensitivity refers to the change in the final loss induced by parameter perturbations of the same scale. As shown in Figure~\ref{fig:perturbation_loss_change}, the dominant layer exhibits a strong perturbation-induced loss response, consistent with its effectiveness in layer-wise ZO fine-tuning. In contrast, most other layers induce only minor loss change, which explains their limited contribution to fine-tuning performance.


However, the magnitude of the change in loss alone does not fully explain the dominant-layer phenomenon. For example, Figure~\ref{fig:perturbation_loss_change} also shows that some later layers, such as layer 30, can produce noticeable loss changes under perturbation. Nevertheless, these layers do not achieve comparable ZO fine-tuning performance as shown in Figure~\ref{fig:mezo_layerwise_three_datasets}. This suggests that a large perturbation-induced loss change alone does not ensure effective ZO fine-tuning. This observation motivates considering both perturbation sensitivity and the position at which perturbations enter the residual stream. 

\begin{figure*}[t]
    \centering
\includegraphics[width=\textwidth]{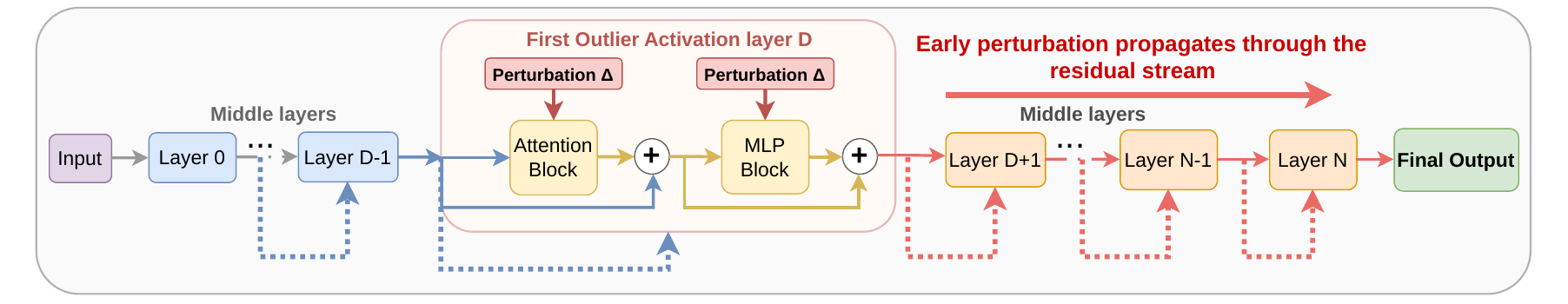}
    \caption{Schematic view of how ZO perturbations are injected at the first outlier-activation layer and propagate through later decoding layers via the residual stream.}
    \label{fig:zo_residual_perturbation}
\end{figure*}

\begin{figure*}[htbp]
    \centering
\includegraphics[width=\textwidth]{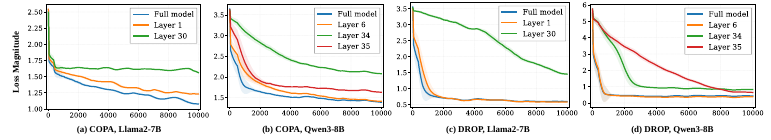}
    \caption{Training loss curve comparison. In both LLaMA2-7B and Qwen3-8B, the dominant layer reduces training loss more effectively than later activation outlier layers.}
    \label{fig:zo_train_loss_comparison}
\end{figure*}

In contrast, due to residual connections between layers in LLMs, the output activation of an earlier layer can affect remaining layers and accumulate into the final hidden activation, as illustrated in Figure~\ref{fig:zo_residual_perturbation}. This effect is especially important for the dominant layer because it aligns with the first activation-outlier layer, whose output contains extremely large-magnitude activations. Once the perturbation effect from the dominant layer enters the following layers, it can affect their hidden activations and continue to accumulate toward the final activation. Therefore, the resulting loss change is not only large in magnitude, but also comes from a propagated effect across all remaining layers. This propagation helps explain why perturbations at the dominant layer provide an effective signal for ZO fine-tuning.
Figure~\ref{fig:zo_train_loss_comparison} further supports this explanation from the training trajectory. Across both LLaMA2-7B and Qwen3-8B, the dominant layer reduces training loss faster than later outlier layers and follows a trajectory closer to full-model fine-tuning. Although later layers also contain activation outliers, their loss decreases much more slowly, suggesting that activation outliers alone are insufficient; early residual-stream propagation is also important for effective ZO fine-tuning.

%% file: sections/experiments.tex
\section{Experimental Validation of Dominant Layer in ZO Fine-tuning}
\subsection{Experimental Setting}
\paragraph{Models and Datasets.} To evaluate performance, we conduct experiments on LLaMA2-7B \citep{touvron2023llama} and Qwen3-8B \citep{yang2025qwen3} over classification, multiple-choice, and generation tasks used in MeZO \citep{malladi2023fine}, including SST-2 \citep{sst2-2013}, RTE \citep{dagan2005pascal, roy2006pascal, giampiccolo2007pascal, bentivogli2009pascal}, CB \citep{de2019cb}, BoolQ \citep{clark2019boolq}, WSC \citep{levesque2012winograd}, MultiRC \citep{khashabi2018multirc}, COPA \citep{roemmele2011copa}, SQuAD \citep{rajpurkar2016squad}, and DROP \citep{dua2019drop}. We also evaluate Mistral-7B~\citep{jiang2023mistral} on SST-2, WSC, COPA, and DROP.

\paragraph{Comparison Setup.} 
To assess whether a single dominant layer can capture the useful adaptation achieved by full-model ZO fine-tuning, we first compare dominant-layer ZO against full-model MeZO~\citep{malladi2023fine} and MeZO LoRA, the most common PEFT variant of MeZO. We also include zero-shot inference without fine-tuning and full-model first-order fine-tuning with AdamW as references, allowing us to understand the gap between ZO and FO fine-tuning. In addition, we compare with Sparse-MeZO~\citep{liu2024sparse}, which reduces the perturbation parameters through sparse masking. We also compare with AGZO~\citep{lin2026agzo}, which focuses on improving the perturbation direction. Dominant-layer ZO uses the same ZO-SGD update rule as MeZO, but restricts both perturbations and updates to the identified dominant layer. To ensure a controlled comparison, all methods follow the MeZO training and evaluation protocol. We tune the learning rate for all methods and the perturbation scale for ZO methods. Full hyperparameter ranges, training steps and other implementation details are provided in the Appendix. Unless stated, we use random seed 0 for experiments.

\subsection{Main Results}
\label{sec:main_results}

\begin{table}[htbp]
\centering
\caption{Performance of fine-tuning LLaMA2-7B (with 1000 examples). FT: full fine-tuning.}
\label{tab:mezo-llama2-7b}
\vspace{2pt}

\scriptsize
\setlength{\tabcolsep}{3.5pt}
\renewcommand{\arraystretch}{1.0}

\resizebox{\columnwidth}{!}{%
\begin{tabular}{lcccccccccc}
\toprule
Task
& \textbf{SST-2} & \textbf{RTE} & \textbf{CB} & \textbf{BoolQ} & \textbf{WSC} & \textbf{MultiRC}
& \textbf{COPA} & \textbf{SQuAD} & \textbf{DROP} & \textbf{AVG.} \\
\cmidrule(lr){2-7} \cmidrule(lr){8-8} \cmidrule(lr){9-10} \cmidrule(lr){11-11}
Task Type
& \multicolumn{6}{c}{classification}
& \multicolumn{1}{c}{multiple choice}
& \multicolumn{2}{c}{generation}
& \\
\midrule
Zero-shot w/o finetune     & 58.02 & 61.73 & 32.14 & 66.7 & 36.54 & 45.3 & 81 & 60.43 & 19.73 & 51.29\\
\midrule
First Order Adamw FT          & 95.87 & 84.84 & 85.71 & 86.7 & 71.15 & 82.6 & 86 & 90.68 & 48.74 & 81.37 \\
\midrule
MeZO FT          & 92.32 & 65.34 & 71.43 & 76.7 & 63.46 & 64.4 & 86 & 87.32 & 39.85 & 71.87 \\
MeZO LoRA        & 92.66 & 65.89 & 69.64 & 77.6 & 63.46 & 65.9 & 88 & 86.26 & 40.57 & 72.22  \\
Dominant-layer ZO FT & 90.79 & 67.51 & 69.64 & 76.5 & 64.42 & 65.86 & 87 & 89.2 & 41.05 & 72.44 \\
Dominant-layer ZO LoRA        & 91.02 & 67.44 & 67.86 & 77.8 & 62.5 & 66.52 & 87 & 88.58 & 42.10 & 72.31 \\
\bottomrule
\end{tabular}%
}
\end{table}
\noindent\textbf{Dominant-layer ZO matches or sometimes exceeds full-model ZO performance.} 
Across both model families, restricting ZO updates to a single dominant layer achieves performance comparable to, and sometimes better than, full-model MeZO. On LLaMA2-7B, dominant-layer ZO FT improves over full-model MeZO FT by an average gain of $0.57\%$, with larger task-level gains on RTE ($+2.17\%$). Dominant-layer ZO LoRA also improves over MeZO LoRA by $0.09\%$ on average, with notable gains on SQuAD ($+2.32\%$). The gains are higher on Qwen3-8B: dominant-layer MeZO improves over full-model MeZO by $1.15\%$ on average, with the largest improvements on CB ($+3.57\%$) and SST-2 ($+2.04\%$). Dominant-layer MeZO LoRA similarly improves over MeZO LoRA by $1.12\%$ on average, with strong gains on CB ($+3.57\%$) and COPA ($+2\%$). Section~\ref{sec:additional_validation} presents additional results across random seeds and on Mistral-7B.

\begin{table}[htbp]
\centering
\caption{Performance of fine-tuning Qwen3-8B (with 1000 examples). FT: full fine-tuning.}
\label{tab:mezo-qwne3-8b}
\vspace{2pt}

\scriptsize
\setlength{\tabcolsep}{3.5pt}
\renewcommand{\arraystretch}{1.0}

\resizebox{\columnwidth}{!}{%
\begin{tabular}{lcccccccccc}
\toprule
Task
& \textbf{SST-2} & \textbf{RTE} & \textbf{CB} & \textbf{BoolQ} & \textbf{WSC} & \textbf{MultiRC}
& \textbf{COPA} & \textbf{SQuAD} & \textbf{DROP} & \textbf{AVG} \\
\cmidrule(lr){2-7} \cmidrule(lr){8-8} \cmidrule(lr){9-10} \cmidrule(lr){11-11}
Task type
& \multicolumn{6}{c}{classification}
& \multicolumn{1}{c}{multiple choice}
& \multicolumn{2}{c}{generation}
& \\
\midrule
Zero-shot w/o finetune     & 58.03 & 87 & 82.14 & 78.3 & 70.19 & 76.4 & 82 & 82.97 & 63.55 & 75.62\\
\midrule
First Order Adamw FT          & 95.41 & 92.06 & 94.64 & 89.9 & 78.85 & 90.1 & 89 & 93.65 & 71.63 & 88.36 \\
\midrule
MeZO FT      & 92.11 & 90.25 & 92.86 & 85.0 & 70.19 & 87.2 & 89 & 89.98 & 64.12 & 84.52 \\
MeZO LoRA        & 91.74 & 90.75 & 91.07 & 85.5 & 71.15 & 86.8 & 87 & 90.41 & 64.59 & 84.33 \\
Dominant-layer MeZO & 94.15 & 91.34 & 96.43 & 84.7 & 72.12 & 85.6 & 90 & 90.76 & 65.94 & 85.67 \\
Dominant-layer MeZO LoRA        & 92.89 & 91.53 & 94.64 & 86.9 & 73.08 & 85.49 & 89 & 90.74 & 64.85 &  85.46  \\
\bottomrule
\end{tabular}
}
\end{table}

\noindent\textbf{Comparison with Sparse-MeZO.} 
Sparse-MeZO \citep{liu2024sparse} selects low-magnitude weights and restricts the ZO perturbation to this selected weight subset to make ZO fine-tuning more stable. It constructs a binary mask and replaces the dense MeZO perturbation with a masked perturbation, so only the selected weights are updated. Following the default settings, the smallest magnitude selection threshold is set at 25$\%$ before training, while the mask can be regenerated during training by comparing the current parameters with the fixed threshold. Table~\ref{tab:sparsemezo_selected} compares Sparse-MeZO with our dominant-layer ZO on LLaMA2-7B. Sparse-MeZO improves the nine-task average performance by $0.54\%$ over MeZO FT. Dominant-layer MeZO improves $0.57\%$ over MeZO FT and $0.03\%$ over Sparse-MeZO. This indicates that choosing where to apply ZO at the layer level can be as important as choosing parameters within every tensor.

\begin{table}[htbp]
\centering
\caption{Comparison with Sparse-MeZO on LLaMA2-7B. FT: full fine-tuning.}
\label{tab:sparsemezo_selected}
\scriptsize
\setlength{\tabcolsep}{3.5pt}
\renewcommand{\arraystretch}{1.05}
\resizebox{\linewidth}{!}{%
\begin{tabular}{lcccccccccc}
\toprule
Method & SST-2 & RTE & CB & BoolQ & WSC & MultiRC & COPA & SQuAD & DROP & AVG \\
\midrule
MeZO FT & 92.32 & 65.34 & 71.43 & 76.70 & 63.46 & 64.40 & 86.00 & 87.32 & 39.85 & 71.87 \\
Sparse-MeZO & 91.63 & 67.90 & 71.43 & 77.50 & 63.46 & 63.20 & 86.00 & 88.83 & 41.72 & 72.41 \\
Dominant-layer MeZO & 90.79 & 67.51 & 69.64 & 76.50 & 64.42 & 65.86 & 87.00 & 89.20 & 41.05 & 72.44 \\
\bottomrule
\end{tabular}%
}
\end{table}



\noindent\textbf{Comparison with AGZO.}
AGZO~\citep{lin2026agzo} uses activation-guided low-rank perturbations within linear modules to improve the perturbation direction. Table~\ref{tab:agzo_comparison} compares AGZO with full-model and dominant-layer MeZO on SST-2, WSC, and COPA for LLaMA2-7B and Qwen3-8B. On LLaMA2-7B, dominant-layer MeZO improves the three-task average performance by $0.15\%$ over MeZO FT and $0.96\%$ over AGZO. On Qwen3-8B, the corresponding gains are $1.65\%$ and $1.81\%$. This indicates that choosing where to apply ZO needs to be considered while constraining the perturbation direction.
 
\begin{table}[H]
\centering
\caption{AGZO comparison on SST-2, WSC, and COPA.}
\label{tab:agzo_comparison}
\small
\setlength{\tabcolsep}{5pt}
\renewcommand{\arraystretch}{1.05}
\begin{tabular}{llcccc}
\toprule
Model & Method & SST-2 & WSC & COPA & AVG \\
\midrule
\multirow{3}{*}{LLaMA2-7B} & MeZO FT & 92.32 & 63.46 & 86.00 & 80.59 \\
& AGZO & 91.97 & 65.38 & 82.00 & 79.78 \\
& Layer-1 MeZO & 90.79 & 64.42 & 87.00 & 80.74 \\
\midrule
\multirow{3}{*}{Qwen3-8B} & MeZO FT & 92.11 & 70.19 & 89.00 & 83.77 \\
& AGZO & 94.72 & 72.12 & 84.00 & 83.61 \\
& Layer-6 MeZO & 94.15 & 72.12 & 90.00 & 85.42 \\
\bottomrule
\end{tabular}
\end{table}

\subsection{Consistency Across Random Seeds and Models}
\label{sec:additional_validation}

\noindent\textbf{Results across random seeds.}
To assess whether the comparable performance of dominant-layer MeZO persists across random seeds, we repeat full-model and dominant-layer MeZO with three perturbation seeds (0, 42, and 2026) on SST-2, WSC, COPA, and DROP for LLaMA2-7B and Qwen3-8B, keeping other settings fixed. Table~\ref{tab:three_seed_validation} reports mean $\pm$ standard deviation. Dominant-layer MeZO maintains comparable mean performance on this four-task subset, supporting the observation beyond a single run.
 
\begin{table}[htbp]
\centering
\caption{Results across three random seeds on four tasks. Task scores are mean $\pm$ standard deviation. DROP reports F1; the other tasks report accuracy.}
\label{tab:three_seed_validation}
\footnotesize
\setlength{\tabcolsep}{3.5pt}
\renewcommand{\arraystretch}{1.05}
\begin{tabular}{llccccc}
\toprule
Model & Method & SST-2 & WSC & COPA & DROP & AVG \\
\midrule
\multirow{2}{*}{LLaMA2-7B} &  MeZO FT & $92.43 \pm 0.20$ & $62.50 \pm 1.66$ & $85.33 \pm 0.58$ & $39.04 \pm 0.95$ & 69.83 \\
& Layer-1 MeZO & $90.33 \pm 0.52$ & $63.14 \pm 1.47$ & $86.33 \pm 0.58$ & $40.50 \pm 1.19$ & 70.08 \\
\midrule
\multirow{2}{*}{Qwen3-8B} & MeZO FT & $91.25 \pm 1.66$ & $70.83 \pm 0.55$ & $88.00 \pm 1.00$ & $64.37 \pm 0.88$ & 78.61 \\
& Layer-6 MeZO & $92.70 \pm 2.42$ & $71.47 \pm 0.56$ & $88.67 \pm 1.15$ & $66.11 \pm 0.42$ & 79.74 \\
\bottomrule
\end{tabular}
\end{table}

\noindent\textbf{Results on Mistral-7B.}
To examine whether the dominant-layer observation extends to another model family, we evaluate Mistral-7B on the same four tasks. The same inference-only activation analysis identifies layer 1, which we use across all four tasks. Table~\ref{tab:mistral_validation} shows that layer-1 MeZO outperforms full-model MeZO by 0.76 percentage points on average across the four tasks.

\begin{table}[htbp]
\centering
\caption{Mistral-7B results on four tasks. DROP reports F1; the other tasks report accuracy.}
\label{tab:mistral_validation}
\small
\setlength{\tabcolsep}{5pt}
\renewcommand{\arraystretch}{1.05}
\begin{tabular}{lccccc}
\toprule
Method & SST-2 & WSC & COPA & DROP & AVG \\
\midrule
Zero-shot & 55.62 & 37.50 & 85.00 & 28.70 & 51.71 \\
MeZO FT & 92.66 & 55.77 & 85.00 & 43.73 & 69.29 \\
Layer-1 MeZO & 92.78 & 54.81 & 86.00 & 46.60 & 70.05 \\
\bottomrule
\end{tabular}
\end{table}

\subsection{Training Efficiency}
\label{sec:training_efficiency}
Dominant-layer ZO fine-tuning also improves training efficiency. Each MeZO step consists of full-model forward evaluations, parameter perturbation, and parameter update. Since the perturbed losses still require full-model forward passes, restricting the optimized parameters does not substantially reduce the forward cost. Therefore, the expected efficiency gain comes mainly from the parameter-side operations when using Dominant-layer ZO. For LLaMA2-7B, which has 32 decoding layers, updating only one dominant layer gives an ideal parameter-side reduction of about $32\times$ compared with full-model MeZO. The measured results in Table~\ref{tab:runtime_breakdown_speedup} closely match this expectation: single-layer ZO reduces perturbation time by $27.34$--$31.28\times$ and update time by $31.56$--$32.71\times$ across tasks. Thus, dominant-layer ZO realizes nearly the full theoretical saving for parameter perturbation and update. The end-to-end speedup depends on the fraction  of the forward pass. On short-input tasks, where perturbation and update dominate a larger portion of each step, dominant-layer ZO achieves larger improvements, such as $4.52\times$ on COPA and $2.45\times$ on SST-2. On long-input tasks, the full-model forward pass dominates the runtime, so the total speedup is smaller, such as $1.24\times$ on CB and $1.12\times$ on DROP. This confirms that dominant-layer ZO is most beneficial when parameter-side operations are a significant bottleneck.

We also compare with Sparse-MeZO as an elementwise sparse perturbation baseline. Although Sparse-MeZO keeps fewer than $25\%$ of parameters active in our setting, its perturbation and update speedups are much smaller. This is expected because Sparse-MeZO applies scattered masks within tensors, while dominant-layer ZO preserves dense contiguous tensor operations on a single decoding layer. Overall, dominant-layer ZO provides stronger practical speedups by combining a layer-level reduction in updated parameters with efficient dense computation.

\begin{table}[htbp]
\centering
\caption{Per-step runtime breakdown for SST2, CB, WSC, COPA, and DROP under different ZO settings on LLaMA2-7B. (Speedups) are relative to full-model MeZO for the same task and runtime component. For Sparse-MeZO, we exclude dynamic mask construction time to isolate the cost of the core ZO perturbation and update operations.}
\label{tab:runtime_breakdown_speedup}
\small
\setlength{\tabcolsep}{6pt}
\begin{tabular}{llcccc}
\toprule
Task & Param Range & Forward/step & Perturb/step & Update/step & Total/step \\
\midrule
SST2 & Full model & 0.4431s (1.00x) & 0.4849s (1.00x) & 0.2061s (1.00x) & 1.1341s (1.00x) \\
SST2 & Sparse-MeZO & 0.4454s (0.99x) & 0.3714s (1.31x) & 0.1317s (1.57x) & 0.9485s (1.20x) \\
SST2 & Dominant layer & 0.4411s (1.00x) & 0.0155s (31.28x) & 0.0063s (32.71x) & 0.4629s (2.45x) \\
\midrule
CB & Full model & 2.9190s (1.00x) & 0.4867s (1.00x) & 0.2061s (1.00x) & 3.6118s (1.00x) \\
CB & Sparse-MeZO & 2.8469s (1.03x) & 0.3842s (1.27x) & 0.1365s (1.51x) & 3.3676s (1.07x) \\
CB & Dominant layer & 2.8815s (1.01x) & 0.0178s (27.34x) & 0.0065s (31.71x) & 2.9057s (1.24x) \\
\midrule
WSC & Full model & 1.0724s (1.00x) & 0.4858s (1.00x) & 0.2083s (1.00x) & 1.7665s (1.00x) \\
WSC & Sparse-MeZO & 1.0623s (1.01x) & 0.3775s (1.29x) & 0.1301s (1.60x) & 1.5699s (1.13x) \\
WSC & Dominant layer & 1.0766s (1.00x) & 0.0161s (30.17x) & 0.0066s (31.56x) & 1.0992s (1.61x) \\
\midrule
COPA & Full model & 0.1764s (1.00x) & 0.4856s (1.00x) & 0.2081s (1.00x) & 0.8701s (1.00x) \\
COPA & Sparse-MeZO & 0.1655s (1.07x) & 0.3490s (1.39x) & 0.1242s (1.68x) & 0.6387s (1.36x) \\
COPA & Dominant layer & 0.1696s (1.04x) & 0.0163s (29.79x) & 0.0064s (32.52x) & 0.1923s (4.52x) \\
\midrule
DROP & Full model & 5.8392s (1.00x) & 0.4852s (1.00x) & 0.2084s (1.00x) & 6.5328s (1.00x) \\
DROP & Sparse-MeZO & 5.9472s (0.98x) & 0.3707s (1.31x) & 0.1334s (1.56x) & 6.4512s (1.01x) \\
DROP & Dominant layer & 5.8273s (1.00x) & 0.0158s (30.71x) & 0.0066s (31.58x) & 5.8496s (1.12x) \\
\bottomrule
\end{tabular}
\end{table}

Additional measurements of peak GPU memory, optimizer-state storage, and trainable-parameter storage are reported in Appendix~\ref{app:memory}.

\needspace{8\baselineskip}
\subsection{Ablation Study and Analysis}
\label{sec:ablations}

\paragraph{Does combining layers improve over the dominant layer?}

\begin{wraptable}{r}{0.45\linewidth}
\begingroup
\centering
\scriptsize
\setlength{\tabcolsep}{4pt}
\renewcommand{\arraystretch}{1.05}
\caption{Layer-combination results on LLaMA2-7B. FT: full fine-tuning.}
\label{tab:discussion_layer_combination}
\begin{tabular}{lcc}
\toprule
Setting & SST-2 & COPA \\
\midrule
w/o Finetune & 58.02 & 81 \\
MeZO FT& 92.32 & 86 \\
Dominant-layer MeZO & 90.79 & 87 \\
Dominant Layer + Layer 30 & 91.52 & 86 \\
\bottomrule
\end{tabular}
\par
\endgroup
\end{wraptable}

We next study whether the dominant layer can be further improved by adding another layer for tuning. As an example, we select layer 30, which also shows relatively large loss changes under perturbation. Table~\ref{tab:discussion_layer_combination} shows that simply adding another layer does not reliably improve performance. On SST-2, tuning layer 1 together with layer 30 slightly improves over tuning the dominant layer alone, but the gain remains close to full-model MeZO. On COPA, however, the two-layer setting does not improve over the dominant layer.
These task-dependent changes suggest that the benefit of adding a layer depends on how its updates interact with those of the dominant layer.

\paragraph{Impact of components within the dominant layer.}
\begin{table}[htbp]
\begingroup
\centering
\caption{MeZO component ablations on LLaMA2-7B. All component and channel restrictions refer to the dominant layer (layer 1).}
\label{tab:channel_ablation}
\small
\setlength{\tabcolsep}{8pt}
\renewcommand{\arraystretch}{1.05}
\begin{tabular}{@{}lcc@{}}
\toprule
Setting & WSC & COPA \\
\midrule
Zero-shot & 36.54 & 81 \\
MeZO FT & 63.46 & 86 \\
\midrule
Layer-0 MeZO & 37.5 & 82 \\
Full dominant-layer MeZO & 64.42 & 87 \\
\midrule
Attention-only & 37.5 & 85 \\
MLP-only & 62.5 & 86 \\
MLP top 1\% outlier channels & 62.5 & 83 \\
Dominant layer excluding top 1\% MLP outlier channels & 56.73 & 81 \\
MLP excluding top 1\% MLP outlier channels & 37.5 & 81 \\
\bottomrule
\end{tabular}
\par
\endgroup
\end{table}

To examine the contributions of different components, we first compare attention-only and MLP-only tuning within the dominant layer (Table~\ref{tab:channel_ablation}). MLP-only tuning performs better than attention-only tuning on both tasks.

Motivated by prior studies showing that activation outliers occur in a small number of fixed feature dimensions and are closely related to the MLP down-projection layer~\citep{sun2024massive, an2025systematic}, we further examine the corresponding channels inside the dominant layer. Specifically, since these outlier dimensions mainly connect to the MLP module within a decoder layer, we study whether the dominant-layer advantage is driven by the associated MLP channels. Table~\ref{tab:channel_ablation} shows that these activation-outlier MLP channels play a critical role. Fine-tuning only the dominant-layer MLP recovers most of the full dominant-layer performance, and tuning only the top 1\% activation-outlier MLP channels still preserves much of the gain on WSC. In contrast, removing these channels from MLP tuning causes performance to collapse close to the no-fine-tuning baseline, indicating that they provide a major part of the ZO update signal.
At the same time, the dominant-layer effect cannot be fully reduced to this small channel subset. Removing the top 1\% outlier channels from full dominant-layer tuning significantly reduces performance, especially on COPA, but does not completely remove the gain on WSC. These results suggest that activation-outlier MLP channels serve as high-leverage components for ZO adaptation, while the rest of the dominant layer still provides additional adaptation capacity.

\paragraph{Perturbation scale and retained updates.}
To assess whether non-dominant layers benefit from stronger perturbations, we increase the perturbation scale for representative early, middle, and late layers on LLaMA2-7B WSC. Larger scales do not close their performance gap to the dominant layer (Appendix~\ref{app:perturbation_scale}). We also examine updates learned during full-model MeZO fine-tuning by retaining one layer's updates and restoring all other layers to their pretrained weights. The best layer under this keep-only evaluation varies by task and can differ from the independently trained dominant layer (Appendix~\ref{app:keep_only}).

%% file: sections/conclusion.tex
\section{Conclusion}
We study how the effectiveness of zeroth-order fine-tuning varies across LLM layers and identify a single dominant layer. Across two model families and nine downstream tasks, tuning this layer often matches or exceeds full-model MeZO, while matched first-order fine-tuning shows much weaker layer concentration. We further show that this layer aligns with the first activation-outlier layer, enabling inference-only identification before training.
Our analysis suggests that the dominant layer combines high perturbation sensitivity with an early position in the residual stream, allowing its perturbation to affect many subsequent blocks. This produces a stronger forward-loss signal under ZO, which relies on loss differences rather than backpropagated gradients. Overall, our results show that the effectiveness of ZO fine-tuning varies substantially across layers, motivating methods that account for layer identity and within-layer importance.

\noindent\textbf{Limitation.}
There is still a performance gap between Dominant-layer ZO and first-order fine-tuning methods. Dominant-layer ZO still takes many steps to achieve good performance, which remains a problem for applications. Extending the study to a broader range of model architectures and scales, and combining dominant-layer tuning with other ZO optimizers such as ZO-AdaMM \citep{chen2019zoadamm} and FZOO \citep{dang2025fzoofastzerothorderoptimizer}, remain directions for future work.

%% file: sections/acknowledgments.tex
\begin{ack}
This work was supported in part by the National Science Foundation under Grant Nos. 2505326, 2528767, and 2348376, and by a seed grant from the School of Data Science at the University of North Carolina at Charlotte. The authors declare no competing financial interests. We thank the Department of Computer Science at Texas Christian University for its support.
\end{ack}

%% file: sections/appendix.tex
\section{Experimental Details}

\subsection{Tasks, Models, and Metrics}
Following MeZO \citep{malladi2023fine}, we construct each task split by sampling up to 1000 training examples, 500 validation examples, and 1000 evaluation examples when sufficient data are available. For smaller datasets, we follow the same protocol but reduce the validation split accordingly; in particular, for WSC, CB, and COPA we use a validation set of 100 examples. We evaluate two model families, LLaMA2-7B and Qwen3-8B, over classification, multiple-choice, and generation tasks. The additional Mistral-7B evaluation covers SST-2, WSC, COPA, and DROP under the same MeZO protocol. The Countdown ES extension uses a separate setup, specified in Appendix~\ref{app:es}. ZO fine-tuning uses float16 for all three models, while FO fine-tuning on LLaMA2-7B and Qwen3-8B uses bfloat16. FO experiments are conducted on NVIDIA H200 GPUs. ZO experiments use A6000-48GB GPUs for LLaMA2-7B and NVIDIA H200 GPUs for Qwen3-8B and Mistral-7B.

Unless otherwise stated, we keep the task formatting, data budget, and evaluation protocol fixed across methods so that the comparisons isolate the effect of the optimization method or trainable scope. In particular, the comparisons among full-model ZO, dominant-layer ZO, Sparse-MeZO, FO fine-tuning, and MeZO-LoRA use the same prompt family and validation-based model selection procedure. This shared setup is important for interpreting the layerwise results: differences in performance should be attributed to optimization behavior rather than to prompt or data changes.

\paragraph{Activation-based layer identification.}
We collect layer-output activation magnitudes during inference with the pretrained model and use the maximum-to-mean magnitude ratio described in Section~\ref{sec:layer_identification}. The earliest layer with a ratio above 1000 is selected before fine-tuning, following the empirical outlier convention of \citet{an2025systematic}. Figure~\ref{fig:activation_magnitude_four_plots} shows the output activation statistics on COPA and WSC training samples. The selected indices are layer 1 for LLaMA2-7B, layer 6 for Qwen3-8B, and layer 1 for Mistral-7B; the selected layer is then held fixed across the evaluated tasks for each model. In our experiments, using 1 calibration example, 10 examples, or the entire training set yields the same selected layer for each model. For Mistral-7B, we validate this activation-based choice by comparing layer-1 MeZO with full-model MeZO on four tasks.

\subsection{Hyperparameters}
Tables~\ref{tab:llama2_hparams} and~\ref{tab:qwen3_hparams} summarize the hyperparameter grids used in our LLaMA2-7B and Qwen3-8B experiments. For MeZO-style methods, we use constant learning rates, a fixed perturbation scale and 10k steps, while FO fine-tuning with AdamW follows a separate learning-rate grid over 5 epochs. For all methods, we select the final checkpoint based on the lowest validation loss among checkpoints saved every 2k training steps. Following the setting from Subzero \citep{yu2025subzero}, we use default sparse rate 0.75 for Sparse-MeZO across all datasets. For AGZO on LLaMA2-7B and Qwen3-8B, we use the corresponding model's MeZO hyperparameter settings in Tables~\ref{tab:llama2_hparams} and~\ref{tab:qwen3_hparams}.

For both full-model and dominant-layer MeZO on Mistral-7B, we search learning rates over $\{5\times10^{-8}, 1\times10^{-7}, 5\times10^{-7}, 1\times10^{-6}\}$ and fix the perturbation scale at $\epsilon=0.001$.

\begin{table*}[htbp]
\centering
\small
\setlength{\tabcolsep}{8pt}
\renewcommand{\arraystretch}{1.15}
\caption{The hyperparameter grids used for LLaMA2-7B experiments. All weight decay is set to 0. FO FT uses 5 epochs and MeZO uses 10K steps and constant learning rates. We check validation performance and save the best checkpoint every 2k total training steps.}
\label{tab:llama2_hparams}
\begin{tabular}{lll}
\toprule
\textbf{Experiment} & \textbf{Hyperparameters} & \textbf{Values} \\
\midrule
\multirow{3}{*}{MeZO FT}
& Batch size & 16 \\
& Learning rate & $\{1\mathrm{e}{-7}, 5\mathrm{e}{-7}, 1\mathrm{e}{-6}, 5\mathrm{e}{-6}\}$\\
& $\epsilon$ & $1\mathrm{e}{-3}$ \\
\midrule

\multirow{3}{*}{MeZO Single Layer}
& Batch size & 16 \\
& Learning rate & $\{1\mathrm{e}{-7}, 5\mathrm{e}{-7}, 1\mathrm{e}{-6}, 5\mathrm{e}{-6}\}$\\
& $\epsilon$ & $1\mathrm{e}{-3}$ \\
\midrule

\multirow{4}{*}{MeZO (LoRA)}
& Batch size & 16 \\
& Learning rate & $\{5\mathrm{e}{-6}, 1\mathrm{e}{-5}, 2\mathrm{e}{-5}, 5\mathrm{e}{-5}\}$\\
& $\epsilon$ & $1\mathrm{e}{-3}$ \\
& $(r, \alpha)$ & $(8, 16)$ \\
\midrule

\multirow{4}{*}{Sparse-MeZO}
& Batch size & 16 \\
& Learning rate & $\{5\mathrm{e}{-7}, 1\mathrm{e}{-6}, 2\mathrm{e}{-6}, 5\mathrm{e}{-6}\}$\\
& $\epsilon$ & $1\mathrm{e}{-3}$ \\
& sparse rate & 0.75 \\
\midrule

\multirow{2}{*}{FO FT with Adamw}
& Batch size & 8 \\
& Learning rates & $\{1\mathrm{e}{-5}, 5\mathrm{e}{-5}, 1\mathrm{e}{-4}\}$ \\
\bottomrule
\end{tabular}
\end{table*}

\begin{table*}[htbp]
\centering
\small
\setlength{\tabcolsep}{8pt}
\renewcommand{\arraystretch}{1.15}
\caption{The hyperparameter grids used for Qwen3-8B experiments. All weight decay is set to 0. FO FT uses 5 epochs and MeZO uses 10K steps and constant learning rates. We check validation performance and save the best checkpoint every 2k total training steps.}
\label{tab:qwen3_hparams}
\begin{tabular}{lll}
\toprule
\textbf{Experiment} & \textbf{Hyperparameters} & \textbf{Values} \\
\midrule
\multirow{3}{*}{MeZO FT}
& Batch size & 16 \\
& Learning rate & $\{1\mathrm{e}{-7}, 5\mathrm{e}{-7}, 1\mathrm{e}{-6}, 5\mathrm{e}{-6}\}$\\
& $\epsilon$ & $1\mathrm{e}{-3}$ \\
\midrule

\multirow{3}{*}{MeZO Single Layer}
& Batch size & 16 \\
& Learning rate & $\{1\mathrm{e}{-7}, 5\mathrm{e}{-7}, 1\mathrm{e}{-6}, 5\mathrm{e}{-6}\}$\\
& $\epsilon$ & $1\mathrm{e}{-3}$ \\
\midrule

\multirow{4}{*}{MeZO (LoRA)}
& Batch size & 16 \\
& Learning rate & $\{5\mathrm{e}{-6}, 1\mathrm{e}{-5}, 2\mathrm{e}{-5}, 5\mathrm{e}{-5}\}$\\
& $\epsilon$ & $1\mathrm{e}{-3}$ \\
& $(r, \alpha)$ & $(8, 16)$ \\
\midrule

\multirow{2}{*}{FO FT with Adamw}
& Batch size & 8 \\
& Learning rates & $\{1\mathrm{e}{-5}, 5\mathrm{e}{-5}, 1\mathrm{e}{-4}\}$ \\
\bottomrule
\end{tabular}
\end{table*}

The hyperparameter tables also clarify an important aspect of our comparisons: the dominant-layer advantage is not due to using a more favorable optimization budget for the restricted scope. Instead, dominant-layer ZO inherits essentially the same MeZO optimization protocol as the full-model baseline, differing only in which parameters are perturbed and updated. This makes the dominant-layer results in the main paper a structural finding rather than a consequence of hyperparameter tuning.

\subsection{Prompts}
Table~\ref{tab:prompt_templates} lists the prompt templates used in our experiments. We follow MeZO \citep{malladi2023fine} for templates and keep the prompt template fixed across zero-shot evaluation, FO fine-tuning, and ZO fine-tuning for each task. This consistency is especially important for the layerwise analyses, since it ensures that the observed differences across layers and optimization methods are not confounded by changes in verbalization or answer formatting.

\begin{table}[htbp]
\centering
\scriptsize
\setlength{\tabcolsep}{4pt}
\renewcommand{\arraystretch}{1.08}
\caption{Prompt templates used in our experiments. Task types are classification (cls.), multiple-choice (mch.), and question answering (QA). Prompts are same as MeZO \citep{malladi2023fine}.}
\label{tab:prompt_templates}
\begin{tabularx}{\columnwidth}{@{} l c >{\raggedright\arraybackslash}X @{}}
\toprule
\textbf{Dataset} & \textbf{Type} & \textbf{Prompt} \\
\midrule
SST-2 & cls. & \texttt{<text>} It was terrible/great \\

RTE & cls. & \makecell[l]{\texttt{<premise>} \\ Does this mean that ``\texttt{<hypothesis>}'' is true? Yes or No? \\ Yes/No} \\

CB & cls. & \makecell[l]{Suppose \texttt{<premise>} Can we infer that ``\texttt{<hypothesis>}''? \\ Yes, No, or Maybe? \\ Yes/No/Maybe} \\

BoolQ & cls. & \makecell[l]{\texttt{<passage><question>}? \\ Yes/No} \\

WSC & cls. & \makecell[l]{\texttt{<text>} \\ In the previous sentence, does the pronoun ``\texttt{<span2>}'' refer to \texttt{<span1>}? \\ Yes or No? \\ Yes/No} \\

MultiRC & cls. & \makecell[l]{\texttt{<paragraph>} \\ Question: \texttt{<question>} \\ I found this answer ``\texttt{<answer>}''. Is that correct? Yes or No? \\ Yes/No} \\

COPA & mch. & \texttt{<premise>} so/because \texttt{<candidate>} \\

ReCoRD & mch. & \makecell[l]{\texttt{<passage>} \\ \texttt{<query>.replace("@placeholder", <candidate>)} } \\

SQuAD & QA & \makecell[l]{Title: \texttt{<title>} \\ Context: \texttt{<context>} \\ Question: \texttt{<question>} \\ Answer:} \\

DROP & QA & \makecell[l]{Passage: \texttt{<context>} \\ Question: \texttt{<question>} \\ Answer:} \\
\bottomrule
\end{tabularx}
\end{table}

\section{Full Layer Sweeps and Optimizer Extension}
\label{app:layer_sweeps}
This section reports additional numerical results that complement the figures and summaries in the main paper. The main paper emphasizes representative plots and high-level comparisons, whereas the appendix provides exact per-task and per-layer values to make the dominant-layer phenomenon fully transparent. Together, these tables show that the concentration of useful ZO adaptation is both numerically sharp and substantially stronger than the corresponding layerwise variation under FO fine-tuning.

\subsection{Layer-wise Fine-tuning on Qwen3-8B}

\paragraph{Complete 36-layer sweep.}
We independently fine-tune each of Qwen3-8B's 36 decoder layers under the same MeZO configuration on WSC, COPA, and DROP, covering the three task formats in the main benchmark. All other layers remain frozen in each run. Table~\ref{tab:qwen3_mezo_36layer} reports the complete sweep. The activation-selected layer 6 achieves the highest COPA score (90) and DROP F1 (65.94), and ties for the highest WSC score (72.12) with layers 10, 11, 21, 26, 28, and 29. Thus, the activation-based choice is supported on all three tasks, although WSC alone does not identify a unique best layer. This validation extends the complete layer-wise analysis beyond LLaMA2-7B.
 
\begin{table}[htbp]
\centering
\caption{Complete 36-layer MeZO sweep Results on Qwen3-8B. WSC and COPA report accuracy; DROP reports F1.}
\label{tab:qwen3_mezo_36layer}
\small
\setlength{\tabcolsep}{4.5pt}
\renewcommand{\arraystretch}{1.05}
\begin{tabular}{cccc|cccc}
\toprule
Layer & WSC & COPA & DROP & Layer & WSC & COPA & DROP \\
\midrule
0 & 69.23 & 82 & 60.49 & 18 & 71.15 & 83 & 62.16 \\
1 & 70.19 & 82 & 63.10 & 19 & 70.19 & 82 & 63.54 \\
2 & 68.27 & 82 & 63.85 & 20 & 69.23 & 83 & 63.11 \\
3 & 69.23 & 84 & 60.42 & 21 & 72.12 & 83 & 64.33 \\
4 & 71.15 & 83 & 59.73 & 22 & 70.19 & 83 & 63.91 \\
5 & 69.23 & 83 & 61.82 & 23 & 71.15 & 82 & 62.56 \\
6 & 72.12 & 90 & 65.94 & 24 & 69.23 & 82 & 62.21 \\
7 & 71.15 & 85 & 60.51 & 25 & 71.15 & 82 & 62.28 \\
8 & 71.15 & 82 & 60.10 & 26 & 72.12 & 82 & 62.26 \\
9 & 71.15 & 86 & 60.43 & 27 & 71.15 & 82 & 63.39 \\
10 & 72.12 & 85 & 61.02 & 28 & 72.12 & 82 & 61.80 \\
11 & 72.12 & 83 & 61.02 & 29 & 72.12 & 82 & 62.40 \\
12 & 69.23 & 84 & 59.59 & 30 & 71.15 & 82 & 62.93 \\
13 & 71.15 & 83 & 60.95 & 31 & 71.15 & 82 & 62.21 \\
14 & 69.23 & 83 & 60.86 & 32 & 71.15 & 82 & 61.99 \\
15 & 70.19 & 84 & 59.14 & 33 & 71.15 & 82 & 61.98 \\
16 & 70.19 & 86 & 61.99 & 34 & 71.15 & 83 & 63.40 \\
17 & 71.15 & 87 & 63.56 & 35 & 71.15 & 83 & 59.44 \\
\bottomrule
\end{tabular}
\end{table}

\subsection{Detailed Results of FO, ZO Layerwise Fine-tuning on LLaMA2-7B}

\begin{table*}[htbp]
\centering
\small
\setlength{\tabcolsep}{5pt}
\caption{LLaMA2-7B single-layer FO results across all 32 transformer layers on WSC, COPA, and DROP. WSC and COPA report accuracy (\%), while DROP reports F1 (\%).}
\label{tab:llama2_fo_32layer}
{%
\begin{tabular}{cccc|cccc}
\toprule
\multicolumn{4}{c|}{Layers 0--15} & \multicolumn{4}{c}{Layers 16--31} \\
\cmidrule(lr){1-4}\cmidrule(l){5-8}
Layer & WSC & COPA & DROP F1 & Layer & WSC & COPA & DROP F1 \\
\midrule
0  & 64.42 & 87 & 45.31 & 16 & 60.58 & 88 & 40.29 \\
1  & 68.27 & 84 & 43.87 & 17 & 56.73 & 89 & 40.10 \\
2  & 61.54 & 85 & 46.07 & 18 & 58.65 & 84 & 37.31 \\
3  & 60.58 & 83 & 46.53 & 19 & 58.65 & 81 & 37.58 \\
4  & 67.31 & 86 & 44.86 & 20 & 58.65 & 90 & 36.05 \\
5  & 68.27 & 87 & 46.06 & 21 & 55.77 & 86 & 35.88 \\
6  & 67.31 & 86 & 47.48 & 22 & 58.65 & 83 & 35.61 \\
7  & 67.31 & 84 & 45.87 & 23 & 58.65 & 86 & 35.22 \\
8  & 71.15 & 82 & 45.44 & 24 & 60.58 & 81 & 35.23 \\
9  & 67.31 & 84 & 44.94 & 25 & 59.62 & 84 & 35.33 \\
10 & 66.35 & 85 & 45.86 & 26 & 56.73 & 84 & 35.99 \\
11 & 66.35 & 87 & 44.46 & 27 & 53.85 & 80 & 36.13 \\
12 & 63.46 & 87 & 45.04 & 28 & 61.54 & 81 & 35.93 \\
13 & 57.69 & 84 & 43.46 & 29 & 61.54 & 86 & 35.18 \\
14 & 62.50 & 84 & 42.44 & 30 & 60.58 & 83 & 34.26 \\
15 & 64.42 & 89 & 40.20 & 31 & 59.62 & 81 & 33.88 \\
\midrule
\multicolumn{8}{l}{\textbf{Full model w/o finetune:} WSC = 36.54, COPA = 81, DROP F1 = 19.73.} \\
\multicolumn{8}{l}{\textbf{FO full model finetune:} WSC = 71.15, COPA = 86, DROP F1 = 48.74.} \\
 
\bottomrule
\end{tabular}%
}
\end{table*}
We next provide the full layerwise FO results for LLaMA2-7B on three representative tasks. These exact values complement the main-text FO figure and make the contrast with ZO more explicit.

Table~\ref{tab:llama2_fo_32layer} confirms that FO fine-tuning is heterogeneous across layers, but the pattern is relatively distributed. Many early and middle layers provide substantial gains over the zero-shot baseline, and no single layer consistently accounts for the full-model FO performance across WSC, COPA, and DROP. Although some layers stand out on individual tasks, the overall FO picture is broad rather than sharply concentrated. This numerical pattern supports the main-text claim that the dominant-layer bottleneck is not a generic property of all fine-tuning, but is much sharper under forward-only ZO.

\begin{table*}[htbp]
\centering
\small
\setlength{\tabcolsep}{5pt}
\caption{LLaMA2-7B single-layer MeZO results across all 32 transformer layers on WSC, COPA, and DROP. WSC and COPA report accuracy (\%), while DROP reports F1 (\%).}
\label{tab:llama2_mezo_32layer}
{%
\begin{tabular}{cccc|cccc}
\toprule
\multicolumn{4}{c|}{Layers 0--15} & \multicolumn{4}{c}{Layers 16--31} \\
\cmidrule(lr){1-4}\cmidrule(l){5-8}
Layer & WSC & COPA & DROP F1 & Layer & WSC & COPA & DROP F1 \\
\midrule
0  & 37.5 & 82 & 36.95 & 16 & 48.08 & 81 & 21.24 \\
1  & 64.42 & 87 & 41.05 & 17 & 45.19 & 81 & 20.46 \\
2  & 37.5 & 81 & 33.08 & 18 & 45.19 & 81 & 20.38 \\
3  & 41.35 & 81 & 32.47 & 19 & 39.42 & 81 & 19.97 \\
4  & 47.12 & 81 & 34.16 & 20 & 36.54 & 81 & 20.30 \\
5  & 43.27 & 81 & 33.91 & 21 & 36.54 & 81 & 19.67 \\
6  & 43.27 & 81 & 33.71 & 22 & 37.5 & 81 & 19.75 \\
7  & 45.19 & 82 & 33.00 & 23 & 36.54 & 81 & 19.86 \\
8  & 43.27 & 82 & 33.41 & 24 & 37.5 & 81 & 19.70 \\
9  & 44.23 & 81 & 33.61 & 25 & 37.5 & 81 & 19.60\\
10 & 43.27 & 81 & 31.85 & 26 & 36.54 & 81 & 19.73 \\
11 & 43.27 & 81 & 30.69 & 27 & 37.5 & 81 & 19.74 \\
12 & 43.27 & 81 & 30.28 & 28 & 39.42 & 81 & 19.51 \\
13 & 43.27 & 81 & 29.44 & 29 & 45.19 & 81 & 19.71 \\
14 & 45.19 & 81 & 27.45 & 30 & 53.85 & 81 & 21.45 \\
15 & 45.19 & 81 & 27.00 & 31 & 59.62 & 81 & 25.88 \\
\midrule
\multicolumn{8}{l}{\textbf{Full model w/o finetune:} WSC = 36.54, COPA = 81, DROP F1 = 19.73.} \\
\multicolumn{8}{l}{\textbf{MeZO full model finetune:} WSC = 63.5, COPA = 86, DROP F1 = 39.85.} \\
\bottomrule
\end{tabular}%
}
\end{table*}

Table~\ref{tab:llama2_mezo_32layer} reports the corresponding single-layer MeZO ablations on LLaMA2-7B. These exact values provide the numerical counterpart to the main-text layerwise MeZO figure. The contrast with FO is obvious. In Table~\ref{tab:llama2_mezo_32layer}, layer 1 clearly emerges as the dominant layer across WSC, COPA, and DROP, recovering most of the gain of full-model MeZO and even slightly exceeding it on DROP. By comparison, most other layers remain far closer to the zero-shot baseline, especially on COPA and in the later layers on DROP. Although a weaker late-layer rise appears on some tasks, layer 1 remains the strongest independently optimized layer across these three tasks. This comparison concerns independently trained layers and does not locate all useful parameter changes within a full-model MeZO checkpoint.

\subsection{Extension to Evolution Strategies}
\label{app:es}
\paragraph{Countdown evaluation.}
To examine whether the layer-restriction observation extends beyond MeZO's SPSA-based fine-tuning, we evaluate Evolution Strategies (ES)~\citep{qiu2025evolution} on Countdown with Qwen3-8B. We compare full-model ES against three single-layer settings: layer 6, the first activation-outlier layer; layer 34, a late activation-outlier layer; and layer 0, the first decoder layer. All four settings use a population size of 16, 16 training prompts per population member, 200 optimization iterations, perturbation scale $\sigma=0.001$, and learning rate $\alpha=0.0005$. We use greedy decoding and evaluate accuracy on the full 2,000-example test set.

Table~\ref{tab:es_countdown} shows that layer-6 ES reaches 52.35\% accuracy, compared with 58.70\% for full-model ES and 29.15\% without updating the model. It recovers most of the improvement over the initial model and outperforms the other tested single-layer settings: 35.25\% for layer 0 and 32.00\% for layer 34. This provides evidence for the usefulness of the first outlier layer in an additional forward-only optimization setting. Layer-6 ES still falls short of full-model ES, and this experiment covers one model and one task rather than the full MeZO benchmark.
 
\begin{table}[htbp]
\centering
\caption{Evolution Strategies on Countdown with Qwen3-8B. Accuracy is evaluated on the full 2,000-example test set.}
\label{tab:es_countdown}
\small
\setlength{\tabcolsep}{5pt}
\renewcommand{\arraystretch}{1.05}
\begin{tabular}{lc}
\toprule
Trainable scope & Accuracy (\%) \\
\midrule
No update & 29.15 \\
Full model & 58.70 \\
Layer 6 only & 52.35 \\
Layer 0 only & 35.25 \\
Layer 34 only & 32.00 \\
\bottomrule
\end{tabular}
\end{table}

\section{Controlled Ablations}
\label{app:controlled_ablations}

\subsection{Layer Combinations}
We compare tuning layer 1 alone with jointly tuning layers 1 and 30 on LLaMA2-7B. Layer 30 is included because it also produces a noticeable loss change under perturbation. Table~\ref{tab:discussion_layer_combination} shows a mixed result: the combination improves SST-2 from 90.79 to 91.52, while COPA decreases from 87 to 86. Adding a layer with high perturbation sensitivity therefore does not consistently improve on the dominant layer. The perturbation sensitivity and early residual-stream placement discussed in Section~\ref{sec:mechanism} characterize an individual perturbation location. Layer 1 already combines these factors; adding layer 30 does not change where its perturbation enters the network or extend its downstream propagation path.

\needspace{14\baselineskip}
\subsection{Perturbation-Scale Control}
\label{app:perturbation_scale}
We test whether larger parameter perturbations can improve representative non-dominant layers on LLaMA2-7B WSC. The selected layers are 0, 15, and 30, spanning early, middle, and late positions. Table~\ref{tab:epsilon_control} includes the original $\epsilon=0.001$ and three larger values, $0.005$, $0.01$, and $0.05$. The best score for layer 0 rises to 39.50 at $0.005$, while the original scale gives the best scores for layers 15 and 30 (45.19 and 53.85). None of the tested scales closes the gap to layer 1 at 64.42.

Changing $\epsilon$ increases the magnitude of the applied perturbation, but does not change the layer's location or its downstream computation. The observed degradation at larger scales is also consistent with a less local finite-difference approximation. These observations motivate distinguishing perturbation magnitude from layer-wise sensitivity; they do not show that every non-dominant layer or model has been exhaustively tuned.
 
\begin{table}[htbp]
\centering
\caption{Perturbation-scale control on LLaMA2-7B WSC. The grid contains the original $\epsilon=0.001$ and three larger scales; scores report accuracy.}
\label{tab:epsilon_control}
\small
\setlength{\tabcolsep}{5pt}
\renewcommand{\arraystretch}{1.05}
\begin{tabular}{lcccc}
\toprule
Layer & $\epsilon=0.001$ & $\epsilon=0.005$ & $\epsilon=0.01$ & $\epsilon=0.05$ \\
\midrule
0 & 37.5 & 39.5 & 37.5 & 35.58 \\
15 & 45.19 & 44.23 & 44.23 & 37.5 \\
30 & 53.85 & 48.08 & 47.12 & 43.27 \\
\bottomrule
\end{tabular}
\end{table}

\subsection{Post-hoc Retained Updates}
\label{app:keep_only}
Starting from each full-model MeZO checkpoint, we retain the learned update for one decoder layer and restore all other layers to their pretrained weights. We repeat this operation for every decoder layer. Table~\ref{tab:keep_only} summarizes the zero-shot and full-model references, the score obtained by retaining the independently identified dominant layer's update, and the best score across the keep-only sweep with its layer index.

Retaining the dominant layer is best on LLaMA2-7B SST-2, COPA, and DROP, and Qwen3-8B SST-2 and COPA. It is not always best: LLaMA2-7B WSC favors layer 7 (39.42 versus 38.46 for layer 1), Qwen3-8B WSC favors layer 3 (72.12 versus 70.19 for layer 6), and Qwen3-8B DROP favors layer 14 (64.11 versus 63.74 for layer 6). Retaining a single update also need not recover the performance of the full checkpoint, as illustrated by LLaMA2-7B WSC and DROP.

This experiment measures a different property from independent layer-restricted training. In the latter, a layer is optimized while the rest of the model stays fixed throughout training. During full-model MeZO, all layers are instead perturbed and updated jointly using the same scalar loss difference. Their learned changes can depend on changes elsewhere in the model. Post-hoc restoration creates a hybrid of one fine-tuned layer and pretrained layers that was not optimized as a configuration during training. Removing the other updates may break complementary or compensatory changes across layers. This is a possible explanation for the difference between the two experiments, rather than a direct measurement of cross-layer coordination. The keep-only results consequently limit claims about the concentration of useful deltas within full-model checkpoints, while leaving the effectiveness of independent dominant-layer training as a separate empirical observation.
 
\begin{table}[htbp]
\centering
\caption{Post-hoc keep-only results after full-model MeZO training. Each retained-update model restores all other layers to pretrained weights. DROP reports F1; the other tasks report accuracy. Layer indices accompany retained-update scores.}
\label{tab:keep_only}
\footnotesize
\setlength{\tabcolsep}{4.5pt}
\renewcommand{\arraystretch}{1.05}
\begin{tabular}{llcccc}
\toprule
Model & Task & Zero-shot & Full MeZO & Dominant retained & Best retained \\
\midrule
\multirow{4}{*}{LLaMA2-7B} & SST-2 & 58.02 & 92.32 & L1: 89.33 & L1: 89.33 \\
& WSC & 36.54 & 63.46 & L1: 38.46 & L7: 39.42 \\
& COPA & 81.00 & 86.00 & L1: 86.00 & L1: 86.00 \\
& DROP & 19.73 & 39.85 & L1: 31.15 & L1: 31.15 \\
\midrule
\multirow{4}{*}{Qwen3-8B} & SST-2 & 58.03 & 92.11 & L6: 94.04 & L6: 94.04 \\
& WSC & 70.19 & 70.19 & L6: 70.19 & L3: 72.12 \\
& COPA & 82.00 & 89.00 & L6: 86.00 & L6: 86.00 \\
& DROP & 63.55 & 64.12 & L6: 63.74 & L14: 64.11 \\
\bottomrule
\end{tabular}
\end{table}

\section{Runtime and Memory Measurements}
\label{app:efficiency}
\subsection{Computation Efficiency Analysis}
This subsection provides an analytical and empirical breakdown of the runtime consequences of restricting the trainable scope in ZO fine-tuning. The main text argues that dominant-layer and channel-restricted ZO can preserve much of full-model performance while reducing optimization overhead. Here we make that tradeoff explicit by separating the cost of forward computation from the cost of perturbing and updating the selected parameters.

All reported runtime measurements compare ZO methods on LLaMA2-7B using an NVIDIA A6000 GPU (48 GB).

\paragraph{Setup.}
Consider a decoder-only layer with hidden size $H$, MLP size $I$,
batch size $B$, effective sequence length $n$, and $q$ SPSA directions.
Let $F(B,n)$ denote the cost of one full forward pass through the model.
For a trainable parameter subset $S$, let $|S|$ be the number of parameters
that are actually perturbed and updated.

\paragraph{Step cost.}
One MeZO SPSA step performs three parameter perturbations,
two full forward passes, and one parameter update. Thus its cost can be written as
\begin{equation}
T^{(1)}_{\mathrm{step}}(S)=2F(B,n)+\gamma |S|,
\end{equation}
where $\gamma>0$ absorbs the per-parameter perturbation and update cost.
With $q$ directions,
\begin{equation}
T_{\mathrm{step}}(S)=2qF(B,n)+q\gamma |S|.
\label{eq:step-cost}
\end{equation}

\paragraph{Key observation.}
Equation~\eqref{eq:step-cost} shows that changing the trainable scope does
\emph{not} change the dominant forward computation: full-model, single-layer,
and outlier-only MeZO all still require two full forward passes per direction.
The only difference lies in the perturbation/update term $q\gamma |S|$.

\paragraph{Parameter counts.}
For three scopes considered in this paper,
\begin{equation}
|S_{\mathrm{full}}| = P_{\mathrm{full}},
\end{equation}
\begin{equation}
|S_{\mathrm{layer}}| = P_{\mathrm{layer}} = 4H^2 + 3HI + 2H,
\end{equation}
and for $k$ selected MLP outlier channels in one layer,
\begin{equation}
|S_{\mathrm{out}}| = P_{\mathrm{out}} = 3Hk.
\end{equation}
Therefore,
\begin{align}
T_{\mathrm{full}} &= 2qF + q\gamma P_{\mathrm{full}}, \\
T_{\mathrm{layer}} &= 2qF + q\gamma P_{\mathrm{layer}}, \\
T_{\mathrm{out}} &= 2qF + q\gamma P_{\mathrm{out}}.
\end{align}

\paragraph{Runtime implication.}
Assuming perturbation and update are implemented only on the selected
parameters, the step-time speedups satisfy
\begin{equation}
\frac{T_{\mathrm{full}}}{T_{\mathrm{layer}}}
=
\frac{2F+\gamma P_{\mathrm{full}}}{2F+\gamma P_{\mathrm{layer}}},
\qquad
\frac{T_{\mathrm{layer}}}{T_{\mathrm{out}}}
=
\frac{2F+\gamma P_{\mathrm{layer}}}{2F+\gamma P_{\mathrm{out}}}.
\end{equation}
Hence the achievable end-to-end speedup is always smaller than the raw
parameter-count ratio, and approaches that ratio only when perturbation/update
dominates the forward cost.

This explains why reducing trainable scope can greatly decrease the
perturbation/update overhead, yet may yield only moderate wall-clock speedup
when full-sequence forward passes already dominate total runtime.

\begin{table}[htbp]
\centering
\caption{Per-step runtime breakdown for SST2, CB, WSC, COPA, and DROP under fine-tuning different model parameter ranges on LLaMA2-7B. Sparse-MeZO excludes dynamic mask-construction time.}
\label{tab:runtime_breakdown}
\small
\setlength{\tabcolsep}{6pt}
\begin{tabular}{llcccc}
\toprule
Task & Param Range & Forward/step & Perturb/step & Update/step & Total/step \\
\midrule
SST2 & full model    & 0.4431s (39.07\%)  & 0.4849s (42.76\%) & 0.2061s (18.18\%) & 1.1341s \\
SST2 & single layer   & 0.4411s (95.273\%) & 0.0155s (3.355\%) & 0.0063s (1.372\%) & 0.4629s \\
SST2 & 1\% mlp channel & 0.4421s (99.377\%) & 0.0024s (0.537\%) & 0.0004s (0.086\%) & 0.4449s \\
SST2 & Sparse-MeZO & 0.4454s (46.96\%) & 0.3714s (39.16\%) & 0.1317s (13.89\%) & 0.9485s \\
\midrule
CB & full model    & 2.9190s (80.82\%) & 0.4867s (13.48\%) & 0.2061s (5.7\%)  & 3.6118s \\
CB & single layer   & 2.8815s (99.17\%) & 0.0178s (0.61\%)  & 0.0065s (0.22\%) & 2.9057s \\
CB & 1\% mlp channel & 2.8811s (99.68\%) & 0.0088s (0.3\%)   & 0.0004s (0.02\%) & 2.8903s \\
CB & Sparse-MeZO & 2.8469s (84.54\%) & 0.3842s (11.41\%) & 0.1365s (4.05\%) & 3.3676s \\
\midrule
WSC & full model    & 1.0724s (60.71\%) & 0.4858s (27.50\%) & 0.2083s (11.79\%) & 1.7665s \\
WSC & single layer   & 1.0766s (97.94\%) & 0.0161s (1.46\%)  & 0.0066s (0.60\%)  & 1.0992s \\
WSC & 1\% mlp channel & 1.0732s (99.55\%) & 0.0044s (0.41\%)  & 0.0004s (0.04\%)  & 1.0781s \\
WSC & Sparse-MeZO & 1.0623s (67.67\%) & 0.3775s (24.05\%) & 0.1301s (8.29\%) & 1.5699s \\
\midrule
COPA & full model    & 0.1764s (20.27\%) & 0.4856s (55.81\%) & 0.2081s (23.91\%) & 0.8701s \\
COPA & single layer   & 0.1696s (88.17\%) & 0.0163s (8.49\%)  & 0.0064s (3.34\%)  & 0.1923s \\
COPA & 1\% mlp channel & 0.1691s (96.70\%) & 0.0053s (3.05\%)  & 0.0004s (0.25\%)  & 0.1749s \\
COPA & Sparse-MeZO & 0.1655s (25.91\%) & 0.3490s (54.64\%) & 0.1242s (19.45\%) & 0.6387s \\
\midrule
DROP & full model    & 5.8392s (89.38\%) & 0.4852s (7.43\%)  & 0.2084s (3.19\%)  & 6.5328s \\
DROP & single layer   & 5.8273s (99.62\%) & 0.0158s (0.27\%)  & 0.0066s (0.11\%)  & 5.8496s \\
DROP & 1\% mlp channel & 5.8334s (99.94\%) & 0.0029s (0.05\%)  & 0.0004s (0.01\%)  & 5.8368s \\
DROP & Sparse-MeZO & 5.9472s (92.19\%) & 0.3707s (5.75\%) & 0.1334s (2.07\%) & 6.4512s \\
\bottomrule
\end{tabular}
\end{table}

Table~\ref{tab:runtime_breakdown} grounds the analytical cost model in measured step-time statistics. Across all tasks, restricting the trainable scope leaves the forward-pass cost nearly unchanged, since the full model must still be executed to obtain the loss. The major difference lies in perturbation and update overhead, which shrinks dramatically for single-layer and 1\% MLP-channel settings. Sparse-MeZO falls between these extremes: it reduces perturbation overhead relative to full-model MeZO, but still incurs substantially more optimization-side cost than dominant-layer or channel-restricted tuning. These measurements clarify why scope restriction can deliver substantial end-to-end savings while still falling short of the raw parameter-count reduction when forward evaluation dominates runtime.

On COPA, perturbation and update account for about 80\% of a full-model MeZO step, explaining the substantial end-to-end speedup from restricting these operations to one layer. On DROP, forward computation accounts for about 89.4\% of the step, leaving less scope for parameter-side savings.

\needspace{20\baselineskip}
\subsection{GPU Memory and Parameter Storage}
\label{app:memory}
\paragraph{Peak GPU memory.}
Table~\ref{tab:gpu_memory} separates the number of trainable parameters, optimizer-state storage, and peak GPU memory. Restricting MeZO to the selected layer reduces the optimized parameter count from 6,738.416M to 202.383M for LLaMA2-7B, and from 8,190.735M to 192.946M for Qwen3-8B. However, the full model remains resident and participates in each forward evaluation, so peak memory changes only from 12.920 to 12.828\,GiB and from 16.481 to 16.013\,GiB, respectively.

Both MeZO scopes have zero reported optimizer-state storage, while FO AdamW uses 25.103\,GiB on LLaMA2-7B and 30.513\,GiB on Qwen3-8B. Its measured peak memory is 50.879 and 63.561\,GiB, respectively. These results distinguish the forward-only memory benefit relative to FO from the relatively small peak-memory difference between the two MeZO scopes.
 
\begin{table}[htbp]
\centering
\caption{Parameter counts and GPU memory measurements. Trainable parameters are reported in millions (M), and optimizer-state and peak GPU memory in GiB.}
\label{tab:gpu_memory}
\small
\setlength{\tabcolsep}{4pt}
\renewcommand{\arraystretch}{1.05}
\begin{tabular}{llrrr}
\toprule
Model & Method & Trainable (M) & Optimizer state & Peak GPU \\
\midrule
\multirow{3}{*}{LLaMA2-7B} & Full FO (AdamW) & 6,738.416 & 25.103 & 50.879 \\
& Full MeZO & 6,738.416 & 0 & 12.920 \\
& Layer-1 MeZO & 202.383 & 0 & 12.828 \\
\midrule
\multirow{3}{*}{Qwen3-8B} & Full FO (AdamW) & 8,190.735 & 30.513 & 63.561 \\
& Full MeZO & 8,190.735 & 0 & 16.481 \\
& Layer-6 MeZO & 192.946 & 0 & 16.013 \\
\bottomrule
\end{tabular}
\end{table}

\paragraph{LoRA parameter storage.}
For LoRA-based MeZO on LLaMA2-7B, Table~\ref{tab:lora_storage} compares adapters distributed throughout the model with adapters restricted to layer 1. The trainable parameter count decreases from 19.988M to 0.625M, and the corresponding storage decreases from 38.125 to 1.191\,MiB. These measurements concern the trainable adapters only. The pretrained model weights and full forward computation are still required.
\begin{table}[htbp]
\centering
\caption{LoRA trainable-parameter storage for LLaMA2-7B.}
\label{tab:lora_storage}
\small
\setlength{\tabcolsep}{5pt}
\renewcommand{\arraystretch}{1.05}
\begin{tabular}{lrr}
\toprule
Scope & Trainable parameters (M) & Storage (MiB) \\
\midrule
Full-model MeZO-LoRA & 19.988 & 38.125 \\
Layer-1 MeZO-LoRA & 0.625 & 1.191 \\
\bottomrule
\end{tabular}
\end{table}